\title{Bayesian Screening: \\ Multi-test Bayesian Optimization Applied to \emph{in silico} Material Screening}
\author{James Hook, Calum Hand, Emma Whitfield}

\documentclass[12pt]{article}
\usepackage{eqnarray,amssymb}
\usepackage{amsmath,amsthm}
\usepackage{pgf}
\usepackage{tikz}
\usetikzlibrary{arrows,chains,matrix,positioning,scopes,shapes}
\usepackage[utf8]{inputenc}
\usepackage{graphicx}

\usepackage{float}
\usepackage{enumitem}
\usepackage{subfigure}
\usepackage{caption}
\usepackage{longtable}
\usepackage{stmaryrd}
\usepackage{hyperref}

\DeclareFontFamily{OT1}{pzc}{}
\DeclareFontShape{OT1}{pzc}{m}{it}{<-> s * [1.10] pzcmi7t}{}
\DeclareMathAlphabet{\mathpzc}{OT1}{pzc}{m}{it}

\usepackage{algorithm}
\usepackage{algpseudocode}
\usepackage{bm}
\usepackage{graphicx}
\usepackage{float}
\usepackage{mathtools}

\DeclareFontFamily{OT1}{pzc}{}
\DeclareFontShape{OT1}{pzc}{m}{it}{<-> s * [1.10] pzcmi7t}{}
\DeclareMathAlphabet{\mathpzc}{OT1}{pzc}{m}{it}

\def\uc#1{\mathcal{#1}}

\def\R{\mathbb{R}}

\def\uc{\mathcal}

\usepackage{bbm}

\usepackage{subfigure}
\usepackage{caption}
\tikzstyle{decision} = [diamond, draw, fill=blue!20, 
    text width=4.5em, text badly centered, node distance=3cm, inner sep=0pt]
\tikzstyle{block} = [rectangle, draw, fill=blue!20, 
    text width=5em, text centered, rounded corners, minimum height=4em]
\tikzstyle{line} = [draw, -latex']
\tikzstyle{cloud} = [draw, ellipse,fill=red!20, node distance=3cm,
    minimum height=2em]

\usepackage{multicol}

\theoremstyle{definition}
\usepackage[hmargin=2cm,vmargin=2cm]{geometry}

\begin{document}
\maketitle

\begin{abstract}
We present new multi-test Bayesian optimization models and algorithms for use in large scale material screening applications. Our screening problems are designed around two tests, one expensive and one cheap. This paper differs from other recent work on multi-test Bayesian optimization through use of a flexible model that allows for complex, non-linear relationships between the cheap and expensive test scores. This additional modeling flexibility is essential in the material screening applications which we describe. We demonstrate the power of our new algorithms on a family of synthetic toy problems as well as on real data from two large scale screening studies. 
\end{abstract}

\section{Introduction}
\label{intro}

In a material (or chemical) screen, the aim is to select from a large number of candidate materials the material or set of materials that perform best in some particular test. This test could be a laboratory experiment or as in our case a computer simulation. Bayesian optimization has previously been suggested as a way to improve the efficiency of such screens, dramatically reducing the number of tests necessary to find the best performing materials. See for example \cite{Frazier2016,Hernandez-Lobato:2017:PDT:3305381.3305533,activenature}. Bayesian optimization is a family of methods for optimizing the output of a black-box function, which work by modeling the input-output relationship of the function as a random variable. Applied to a materials screen, Bayesian optimization works by iteratively selecting new materials for testing whilst simultaneously using the data generated by the previous tests to model the relationship between known properties of the candidate materials and their performance in the test. The model's predictions are then used to guide further experimentation.

There has been recent interest in developing Bayesian optimization methods that can efficiently choose between a range of tests with different cost/accuracy trade offs (in \emph{in silico} material screening the cost is computional time). See \cite{pmlr-v89-song19b} for an introduction. Application examples include tuning the hyperparameters for deep learning models, where a cheap but less accurate test can be made by training on a smaller data set or with fewer iterations \cite{wu2019practical}. Another proposed application is testing a robot control strategy either with a real life experiment or a computer simulation \cite{Kandasamy:2016:GPB:3157096.3157208}. 

Chemical engineers conducting screening studies have access to a huge number of possible tests. These range from calculating basic features of the candidate materials, which can be done in factions of a second, to detailed real life experiments, which might takes weeks or months to prepare. Screening studies typically work by applying a cheaper test to all of the candidates and then selecting some top fraction of performers to move onto a second round and so on. Utilizing high performance computing, this approach allows for high-throughput \emph{in silico} screening of huge databases of materials \cite{naturerevierw}. For example the authors of \cite{mofscreen} apply three rounds of increasingly accurate but expensive computer simulated experiments to find the most promising candidate materials in a database of 137,953 structures. This method is also widely used in screens where all of the test are real life experiments, including in industry \cite{ESGI}. In practise this approach has been shown to work well in many applications but the choice of which tests to apply and how many candidates to pass through each round tends to be carried out in an \emph{ad hoc} fashion without the use of available statistical and AI techniques. 

In this paper we present new algorithms for multi-test Bayesian optimization, with a novel objective and statistical model for application in large scale materials screening problems.


\section{Models for screening and data}

There are $n$ candidates which we index by $1,2,\dots,n$. Each candidate has a vector of features $\bm{x}_{i}\in\R^d$, which is visible to the decision making algorithm throughout the process. Additionally each candidate has a cheap test score $y_{i}^{(C)}\in\R$ and an expensive test score $y_{i}^{(E)}\in\R$, which are initially hidden and only become available after they are tested for. Applying the cheap test costs $c_{C}$ and applying the expensive test costs $c_{E}$. We consider two objectives for the screening problem as follows. 

\smallskip

{\bf Optimization:} Our aim is to use a fixed testing budget $B$ to find the candidate with the highest expensive test score that we can. 

{\bf Mining:} Our aim is to use a fixed testing budget $B$ to find as many of the candidates with expensive test scores in the top $N$ as we can.

\smallskip

Previous work applying Bayesian optimization to material screening has focused on the optimization objective. However, when searching very large computer databases of candidate materials, the chemical engineers that we have spoken to say that they really want to find all of the top performing materials. This can be for many reasons including that some materials in the database will be difficult or impossible to synthesize. Therefore the mining objective is a better model for what the chemical engineers actually want and targeting this objective may result in better performance in practice. We also consider two different models for the hidden test scores as follows.

\smallskip

{\bf Multi-Fidelity Testing Model:} To model the scores, we consider a Gaussian Process $\{f(\bm{x})\}_{\bm{x}\in\R^d}$, then define the cheap and expensive test scores  $\bm{y}^{(C)},\bm{y}^{(E)}\in\R^n$, by
\begin{equation}\label{mf}
y^{(C)}_{i}=f(\bm{x}_{i})+\epsilon_{i}^{(C)}, \quad y^{(E)}_{i}=f(\bm{x}_{i})+\epsilon_{i}^{(E)},
\end{equation}
for $i=1,\dots,n$, where $(\epsilon_{i}^{(C)})_{i=1}^{n}$ are i.i.d. $\uc{N}(0,\sigma_{C}^2)$ and $(\epsilon_{i}^{(E)})_{i=1}^{n}$ are i.i.d. $\uc{N}(0,\sigma_{E}^2)$. 

{\bf Covariate Testing Model:} To model the scores, we consider two Gaussian Process $\{f(\bm{x})\}_{\bm{x}\in\R^d}$ and \\ $\{g(\bm{x},y)\}_{\bm{x}\in\R^d,y\in\R}$, then define the cheap and expensive test scores  $\bm{y}^{(C)},\bm{y}^{(E)}\in\R^n$, by
\begin{equation}\label{ct}
y^{(C)}_{i}=f(\bm{x}_{i})+\epsilon_{i}^{(C)}, \quad y^{(E)}_{i}=g(\bm{x}_{i},y_{i}^{(C)})+\epsilon_{i}^{(E)},
\end{equation}
for $i=1,\dots,n$, where $(\epsilon_{i}^{(C)})_{i=1}^{n}$ are i.i.d. $\uc{N}(0,\sigma_{C}^2)$ and $(\epsilon_{i}^{(E)})_{i=1}^{n}$ are i.i.d. $\uc{N}(0,\sigma_{E}^2)$. 

\smallskip

Previous work on multi-test Bayesian optimization has focused on the multi-fidelity model or on multi-dimensional Gaussian process models that also result in linear relationships between the cheap and expensive test scores. These models results in Gaussian posterior distributions which are easy to work with but are only appropriate in screening problems where the cheap test gives an unbiased estimate of the expensive test or where the cheap test and expensive test are related in a fixed linear way. We are interested in screening problems where the cheap test scores provides useful information about the expensive test scores but via some initially unknown non-linear relationship. In such problems we need the extra flexibility of the covariate testing model. However this flexibility comes at the price of tractability as the resulting posterior distributions are non-Gaussian which makes inference more complicated and expensive.

\subsection{Markov Decision Process (MDP) formulation}\label{MDP}

Assuming any of the four possible objective/model combinations gives rise to a MDP model for the screening problem. See \cite{Sutton:1998:IRL:551283} Chapter 3 for an introduction. The state space of the MDP is given by $S=\big((\R\cup\{?\})^2\big)^n$, where real values represent scores that have been tested for and question marks represent so far unknown test scores. For example, a state $s\in S$, with $s_{i}=(a,?)$, where $a\in\R$, indicates that candidate $i$ has been tested with the cheap test and scored $y_{i}^{(C)}=a$ but that candidate $i$ has not been tested with the expensive test. For a state $s\in S$, define 
\begin{equation}\label{stateprops}
\begin{array}{c}
\begin{array}{cc} I_{\hbox{uu}}=\{i~:~s_{i}=(?,?)\}, & I_{\hbox{tu}}=\{i~:~s_{i}\in \R\times \{?\}\}, \\  
I_{\hbox{ut}}=\{i~:~s_{i}\in \{?\}\times\R\}, & I_{\hbox{tt}}=\{i~:~s_{i}\in \R\times \R\},\end{array}
 \\
\begin{array}{c}
D=\{\bm{y}^{(C)}_{I_{\hbox{tu}}\cup I_{\hbox{tt}}},\bm{y}^{(E)}_{I_{\hbox{ut}}\cup I_{\hbox{tt}}}\},\\ b=B-c_{C}(|I_{\hbox{tu}}|+|I_{\hbox{tt}}|)-c_{E}(|I_{\hbox{ut}}|+|I_{\hbox{tt}}|),\\
 \uc{A}=\{(i,C)~:~i\in I_{\hbox{uu}}\cup I_{\hbox{ut}}\}\cup\{(i,E)~:~i\in I_{\hbox{uu}}\cup I_{\hbox{tu}}\} \\ y_{\max}^{(E)}=\max_{i\in I_{\hbox{ut}}\cup I_{\hbox{tt} }}y_{i}^{(E)}.\end{array}
\end{array}
\end{equation}

From state $s$ the available actions, denoted by the set $\uc{A}$, are to apply the cheap test or expensive test to any candidate that has not taken that test so far. If there is not sufficient budget for any further tests then the MDP is terminated. When an action is taken the MDP transitions to a new state $s\mapsto s'$ by revealing the score of the chosen test. Assuming either of the statistical models for the test scores, this gives rise to random transitions. For example if we choose action $(i,E)$, i.e. to test candidate $i$ with the expensive test, then $(s'_{i})_{2}\sim \rho(y_{i}^{(E)}|D)$ and all other entries of $s'$ are equal to those of $s$ and where conditioning on $D$ mean conditioning on all of the previously tested values for $\bm{y}^{(C)}$ and $\bm{y}^{(E)}$.

When the MDP transitions $s\mapsto s'$ a reward is obtained. The rewards for applying the cheap test are all equal to zero. For the expensive test under the optimization objective the reward is given by 
\begin{equation}\label{EI}
R\big((i,E),s\big)=\max\left\{\left(y_{i}^{(E)}-y_{\max}^{(E)}\right),0\right\},
\end{equation}
i.e. the improvement in the running maximum, and under the mining objective the reward is given by
\begin{equation}\label{miningreward}
R\big((i,E),s\big)=\left\{\begin{array}{cc} 1 & \hbox{if $i\in$ top$_N$,} \\ 0 & \hbox{otherwise,}\end{array}\right.
\end{equation}
where top$_{N}$ is the set of the $N$ top expensive test scoring candidates. Note that the rewards signal will be hidden to the MDP agent in the case of the mining objecting as the agent cannot know for sure which candidates are in top$_{N}$ until it has tested all of them.  

A policy is a (possibly random) mapping $\pi: S\mapsto \uc{A}$ from states to actions that defined a screening strategy. Once we fix a policy $\pi$ the MDP becomes a Markov chain
\begin{equation}\label{markov}
s_{0} \overset{\pi(s_{0})}{\longmapsto} s_{1}\overset{\pi(s_{1})}{\longmapsto} s_{2} \overset{\pi(s_{2})}{\longmapsto} \cdots \overset{\pi(s_{T-1})}{\longmapsto} s_{T},
\end{equation}
where $s_{0}=\underline{?}$ is the initial state where we have no test data, $s_{t+1}$ is the state we transition to after taking action $\pi(s_{t})$ from state $s_{t}$ for $t=0,1,\dots,T-1$ and where $s_{T}$ is the first state at which we do not have sufficient budget to take any further action. 

The screening problems can now be restated as MDP policy optimization problems by
\begin{equation}\label{taget}
\max_{\pi}\mathbb{E}\left[\sum_{t=0}^{T-1}R\big(\pi(s_{t}),s_{t}\big)\right].
\end{equation}
The focus of this paper is to develop algorithmic policies to work with the mining objective and covariate testing model in large scale material screening applications.

\section{Single-Test Bayesian Optimization}\label{signleBO}

 Classical single-test Bayesian optimization works by updating an \emph{acquisition function}, $\alpha(D)\in\R^n$, at each stage and then sampling the candidate that maximizes it. See  \cite{frazier2018tutorial} for an introduction. Note that in the case of single-test Bayesian optimization there are only two sets of candidates of interest: the untested candidates $I_{\hbox{u}}$ and the tested candidates $I_{\hbox{t}}$.  See Algorithm~\ref{alg1}. 
 
 \begin{algorithm}	
	\caption{Bayesian Optimization}\label{alg1}	
	\begin{algorithmic}[1]
		\While{$b> 0$}
		\State $i^{\ast}=\arg\max_{i\in I_{\hbox{u}}}\alpha_{i}(D)$ 
		\State apply test to $i^{\ast}$
		\EndWhile
	\end{algorithmic}
\end{algorithm}

\section{Two-Test Sequential Bayesian Optimization}

We will restrict ourselves to sequential methods in which candidates have to be tested with the cheap test before they can be tested with the expensive test. This restriction means that $I_{\hbox{ut}}$ will always be empty, which avoids some major difficulties in making inferences with the covariate testing model. In many cases this restriction will be forced on us by practical considerations, for example if the cheap test is an intermediate result that must be tested for as part of the expensive test. However there may also be cases where the two tests are not related in this way and in those cases sequential testing may not be optimal. For example if the cheap test score is not useful or not useful enough to justify its cost then the optimal policy would learn to skip this test, but that will not be possible for a sequential method.

All of our proposed two-test sampling methods are implementations of the same high level algorithm. At each stage Algorithm~\ref{alg2} either applies the cheap test to advance a candidate from $I_{\hbox{uu}}$ to $I_{\hbox{tu}}$ or applies the expensive test to advance a candidate from $I_{\hbox{tu}}$ to $I_{\hbox{tt}}$.

Exactly as in single-test Bayesian optimization an acquisition function is updated at each stage to identify candidates for testing. However now a controller decides whether to apply the cheap test to the best candidate from $I_{\hbox{uu}}$ or to apply the expensive test to the best candidate from $I_{\hbox{tu}}$.  Note that unlike in single-test Bayesian optimization the acquisition function depends on all of the available $\bm{y}^{(C)}$ and $\bm{y}^{(E)}$ data and assigns values to candidates in both $I_{\hbox{uu}}$ and $I_{\hbox{tu}}$. Despite these differences we are able to adapt widely used single-test acquisition functions to our setting using almost exactly the same mathematical definitions, we just need to calculate them slightly differently. We will use the following acquisition functions: \newline

\noindent {\bf Two-Test Greedy Expected Improvement:} This acquisition function returns the expected reward for applying the expensive test to candidate $i$ under the optimization objective. 
\begin{equation}\label{a1}
 \alpha_{i}(D) =\mathbb{E}\big[\max(y_{i}^{(E)}-y_{\max}^{(E)},0)\big|D].
\end{equation}

\noindent {\bf Two-Test Greedy Mining:} This acquisition function returns the expected reward for applying the expensive test to candidate $i$ under the mining objective. 
\begin{equation}\label{a2}
 \alpha_{i}(D) =\mathbb{P}[i\in \hbox{top}_{N}|D].
\end{equation}

\noindent {\bf Two-Test Greedy Threshold:} This acquisition can be used to approximate \eqref{a2}.
\begin{equation}\label{a3}
 \alpha_{i}(D) =\mathbb{P}[y_{i}^{(E)}\geq \tau|D].
\end{equation}

\noindent {\bf Two-Test Thompson:} This random acquisition function is obtained by sampling from the posterior of the expensive test scores 
\begin{equation}\label{a4}
\alpha(D)\sim \rho(\bm{y}^{(E)}|D).
\end{equation}

Accurately estimating the greedy mining acquisition function requires a large number of samples and this can be prohibitively expensive. We therefore propose using the greedy threshold acquisition function as an approximation. The threshold $\tau\in\R$ is chosen to be the posterior median of the $y^{(E)}$ score of the $N$th highest scoring candidate. Although this value also needs to be estimated through sampling, it will have a much lower variance than the greedy acquisition function, so can be accurately estimated from a modest number of samples, and does not need to be updated on every iteration. Because the threshold score is based on the absolute value of each candidates score, rather than their ranking, this approximation cuts out a lot of complex dependencies.

If we fix the choice of acquisition function then the MDP screening problem can be restated from the point of view of the controller as follows. From a state $s\in S$, with sufficient budget, there are two actions available:

\begin{enumerate}
    \item Apply the cheap test test to $i_{uu}=\arg\max_{i\in I_{\hbox{uu}}}\alpha_{i}(D)$.
    \item Apply the expensive test to $i_{tu}=\arg\max_{i\in I_{\hbox{tu}}}\alpha_{i}(D)$.
\end{enumerate}

\begin{algorithm}
	\caption{Sequential Bayesian Optimization}\label{alg2}	
	\begin{algorithmic}[1]
		\While{$b>0$}
		\State $i_{uu}=\arg\max_{i\in I_{\hbox{uu}}}\alpha_{i}(D)$ 
		\State $i_{tu}=\arg\max_{i\in I_{\hbox{tu}}}\alpha_{i}(D)$ 
		\If{ controller chooses action 1}
		\State apply cheap test to $i_{uu}$
		\Else
		\State apply expensive test to $i_{tu}$
		\EndIf
		\EndWhile
	\end{algorithmic}
\end{algorithm}

\subsection{Sequential Greedy}

The Sequential Greedy (SG) method combines a greedy acquisition function with a greedy controller that chooses which test to apply from a state $s\in S$ by comparing the expected reward to cost ratio of two different sequences of actions. 

\begin{enumerate}
    \item Apply the cheap test to $i_{uu}$ and then apply the expensive test to whichever is the most promising of $i_{tu}$ and $i_{uu}$ given the new data from the cheap test.
    \item Apply the expensive test to $i_{tu}$.
\end{enumerate}

The greedy controller chooses action 1 whenever
\begin{align}\label{greedydec}
\nonumber \frac{\mathbb{E}\left[\max\big(\alpha_{i(tu)}(D,y^{(C)}_{i(uu)}),\alpha_{i(uu)}(D,y^{(C)}_{i(uu)})\big)~|~D \right]}{c_{C}+c_{E}}  >\frac{\alpha_{i(tu)}(D)}{c_{E}}
\end{align}
where $\alpha$ is one of the greedy  acquisition functions. See Algorithm~4.1 in the supplementary material.

\subsection{Sequential Thompson with Random Controller}

The Sequential Thompson with Random controller (STR) method combines the Thompson sampling acquisition function with a random controller. The controller chooses between action 1 and action 2 independently and randomly at each stage according to a probability distribution which is chosen as a parameter of the method. One possibility is to set the probability taking action 1 by
\begin{equation}\label{thompsontune}
    p_{1}=\frac{c_{E}+2c_{C}}{c_{E}+3c_{C}}
\end{equation}
which is chosen so that roughly half of the budget is spent on applying the cheap test to candidates which are never tested with the expensive test. See Algorithm 4.2 in the supplementary material.

\begin{figure*}[t!]
\begin{center}
\subfigure[$\theta=0$]{\includegraphics[scale=0.4]{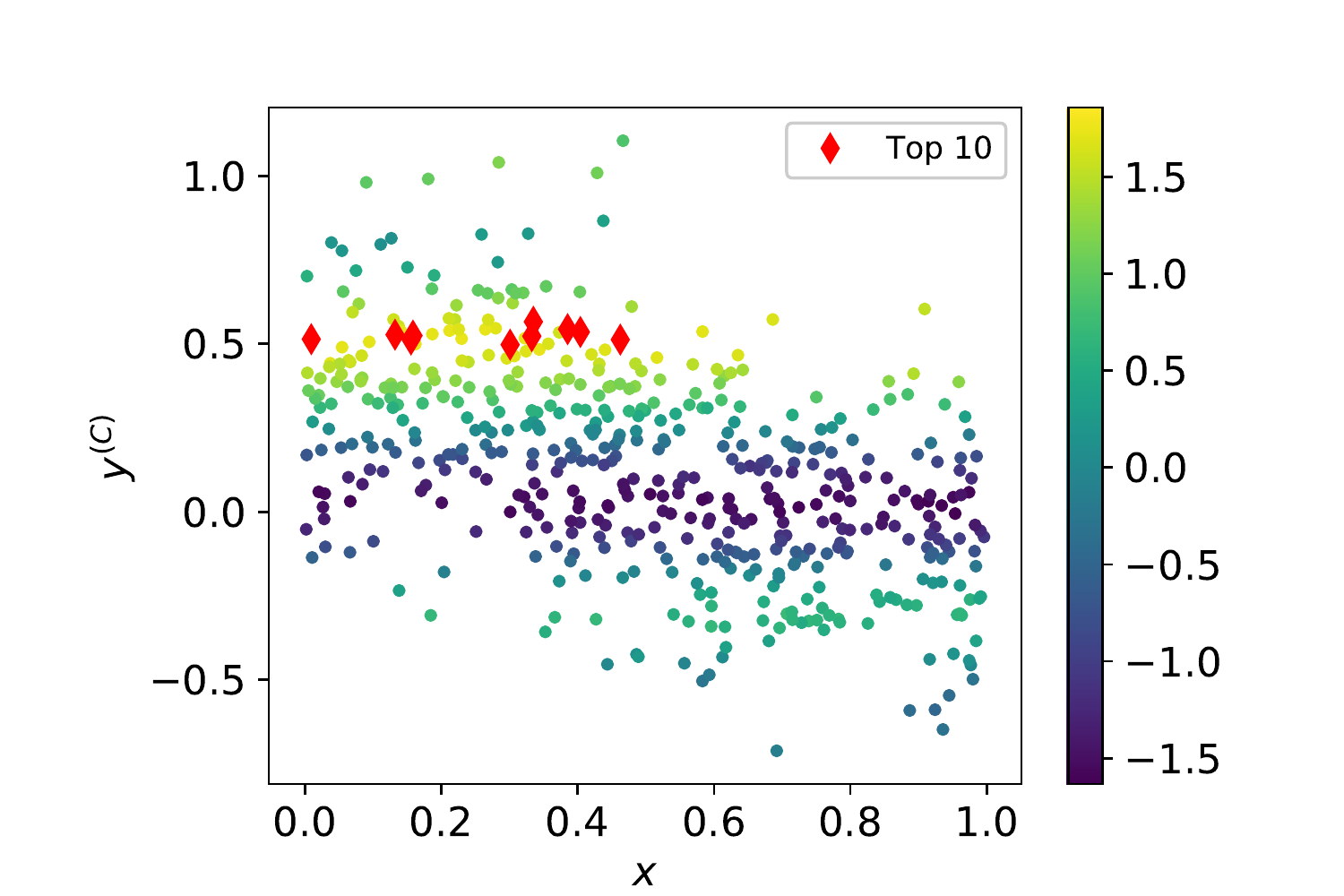}}\subfigure[$\theta=\pi/4$]{\includegraphics[scale=0.4]{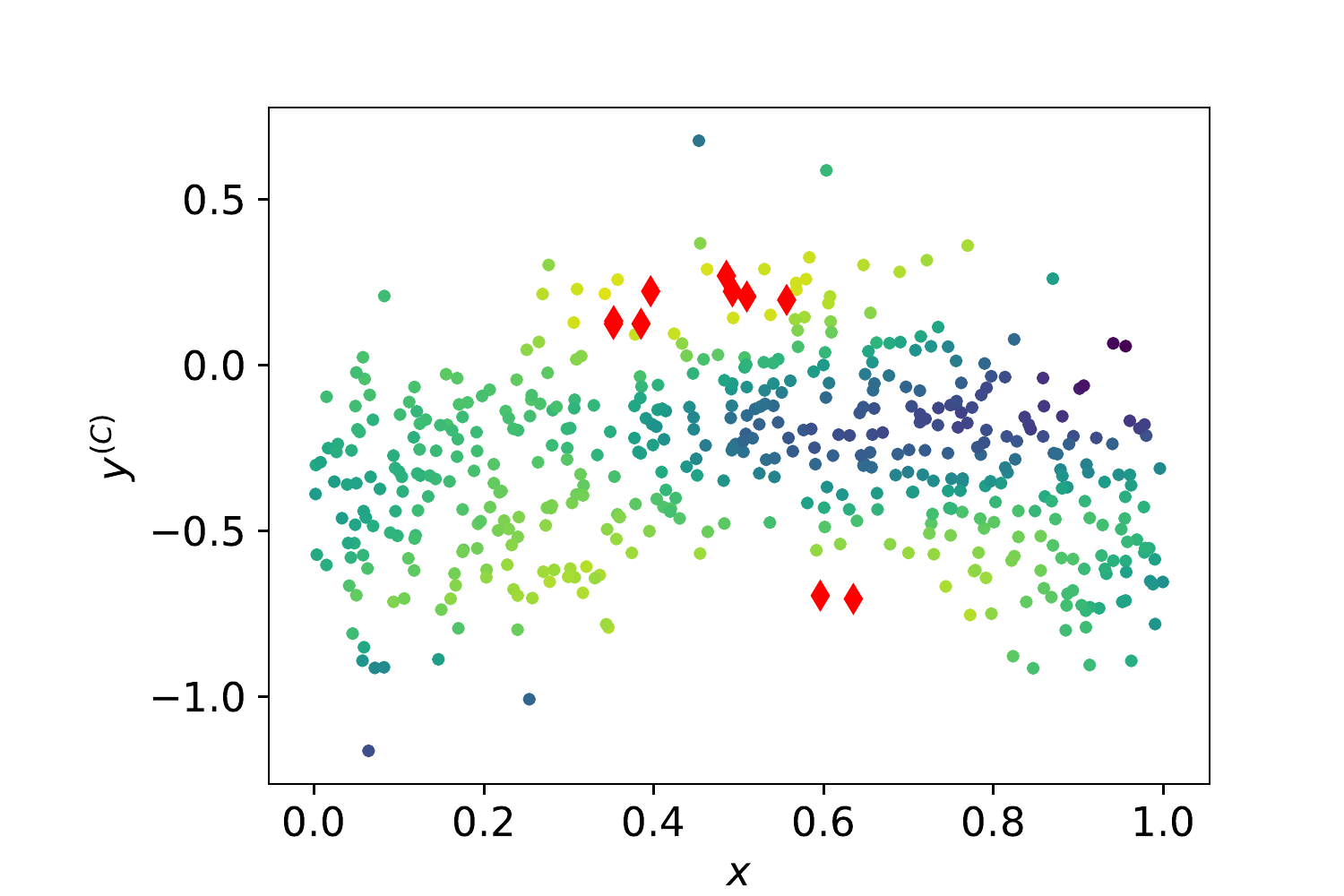}}\subfigure[$\theta=\pi/2$]{\includegraphics[scale=0.4]{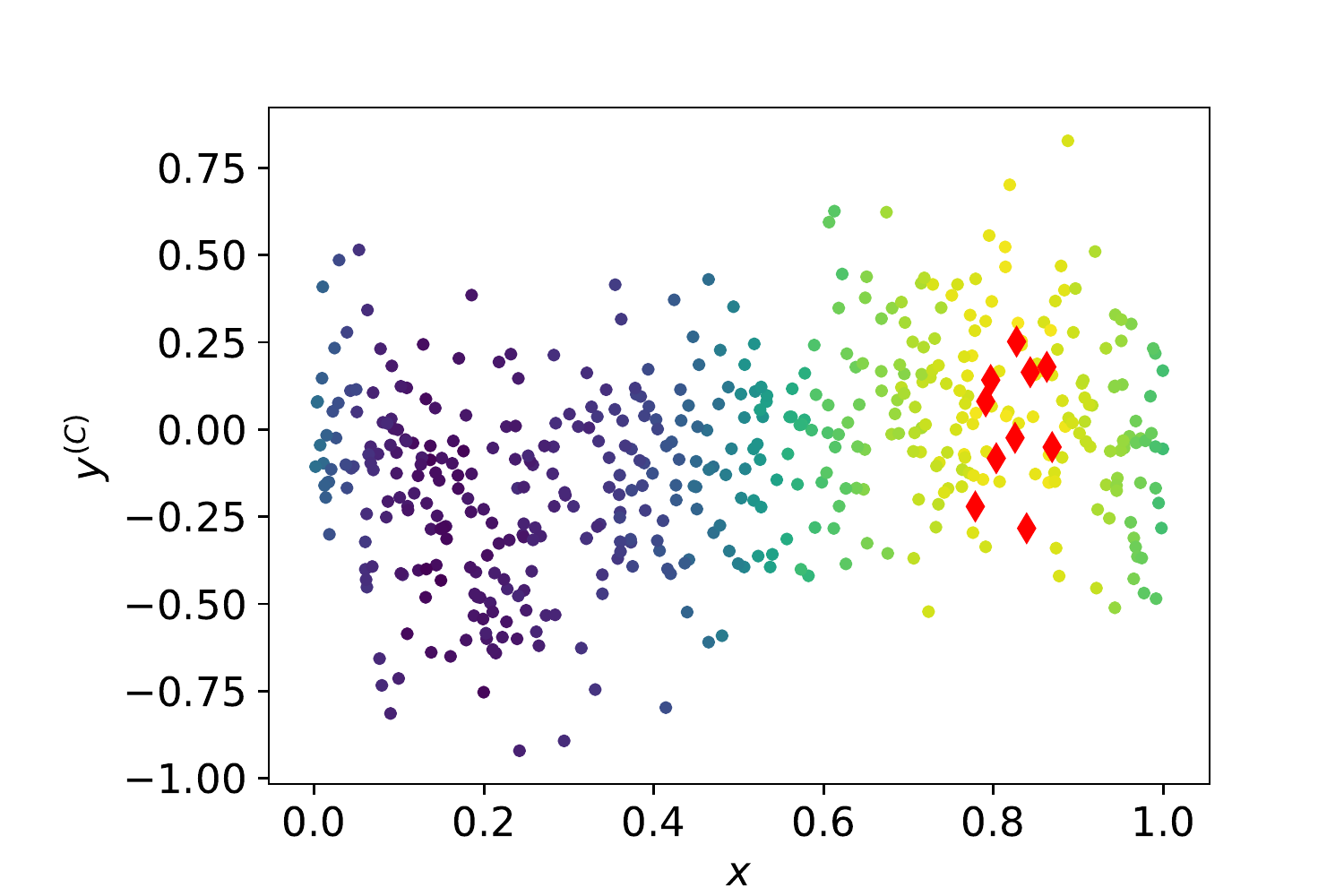}} 
\caption{Example synthetic test data described in Section~\ref{problemegg} with top 10 candidates highlighted.}\label{cofmoffig}
\end{center}
\end{figure*}

\section{Asynchronous Parallel Implementation}

In large scale screening applications it is essential that the sampling method can work efficiently with a large number of asynchronous parallel workers. In \cite{kandasamy18a} the authors show that single-test Thompson sampling is well suited to asynchronous parallel implementation.  In \cite{fb} the authors adapt  single-test expected improvement to this stetting by conditioning on and then marginalizing out the scores of any candidates that are currently being tested. The same adaptation could be applied to two-test expected improvement but unfortunately their method would not make any difference to the two-test greedy threshold acquisition function.

In all of the experiments in this paper we simulate the behaviour of $w\in\mathbb{N}$ asynchronous parallel workers. The times taken to carry out the cheap tests are i.i.d. samples from $U[c_{C}/2,3c_{C}/2]$ and the expensive tests i.i.d. samples from $U[c_{E}/2,3c_{E}/2]$. Each worker uses the sampling algorithm to choose an available action, carries out the associated test and then as soon as it finishes immediately chooses a new action to start.

\section{Test on synthetic data}\label{problemegg}

In this section we test our algorithms in simulated screens on synthetic data. Each problem is generated as follows. We set $n=500$, then sample $\bm{x}\in\R^n$ with i.i.d. uniform $[0,1]$ entries, then we sample $\bm{y}^{(C)}\in\R^n$ from $\uc{N}(\underline{0},\Sigma^{(C)})$, where
\begin{equation}\label{zsynth}
\big(\Sigma^{(C)}\big)_{ij}=0.25^2\exp\left(\frac{-(x_{i}-x_{j})}{2\times 0.25^2}\right)+0.25^2\delta_{ij},
\end{equation}
for $i,j=1,\dots,n$, then we sample $\bm{y}^{(E)}\in\R^n$ from $\uc{N}(\underline{0},\Sigma^{(E)})$, where
\tiny
\begin{equation}\label{ysynth}
\big(\Sigma^{(E)}\big)_{ij}=\exp\left(-\frac{(x_{i}-x_{j})}{2\times 0.25^2}\sin^2(\theta)-\frac{(y^{(C)}_{i}-y^{(C)}_{j})}{2\times 0.25^2}\cos^2(\theta)\right)+0.05^2\delta_{ij},
\end{equation}
\normalsize
for $i,j=1,\dots,n$, where $\theta\in[0,\pi/2]$. 

 Note that in these tests, the sampling algorithms will have full knowledge of the generative model including all of the hyperparameter values. The hyperparameter $\theta$ varies the expensive score's length scales with respect to the $x$ values, which are visible to the algorithm throughout with no cost, and the $y^{(C)}$ values, which need to be tested for to be revealed. The cheap test therefore provides more useful information for smaller values of $\theta$. See Figure~\ref{cofmoffig}.


\begin{figure*}[t!]
\begin{center}
\subfigure[]{\includegraphics[scale=0.4]{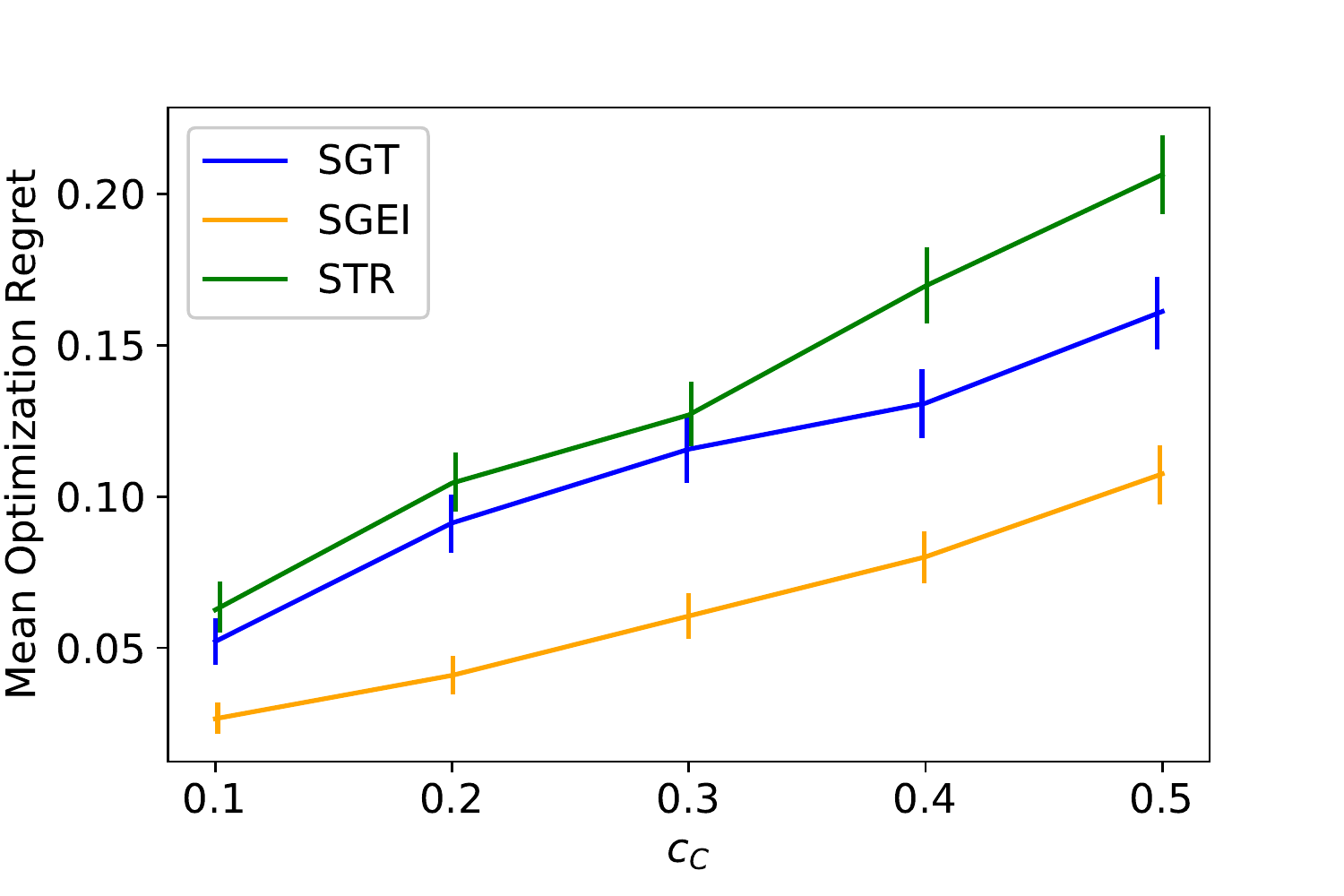}}\subfigure[]{\includegraphics[scale=0.4]{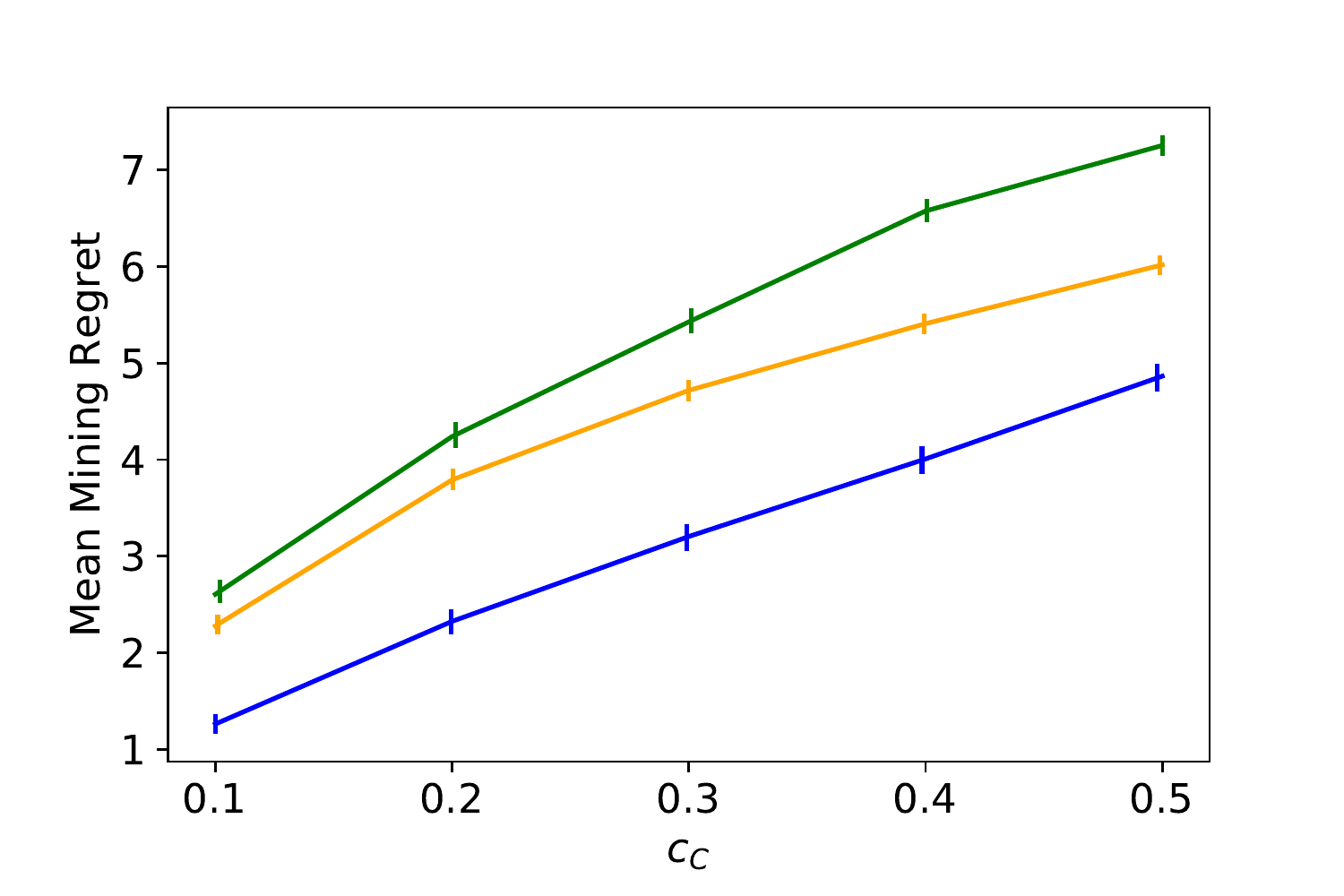}}\subfigure[]{\includegraphics[scale=0.4]{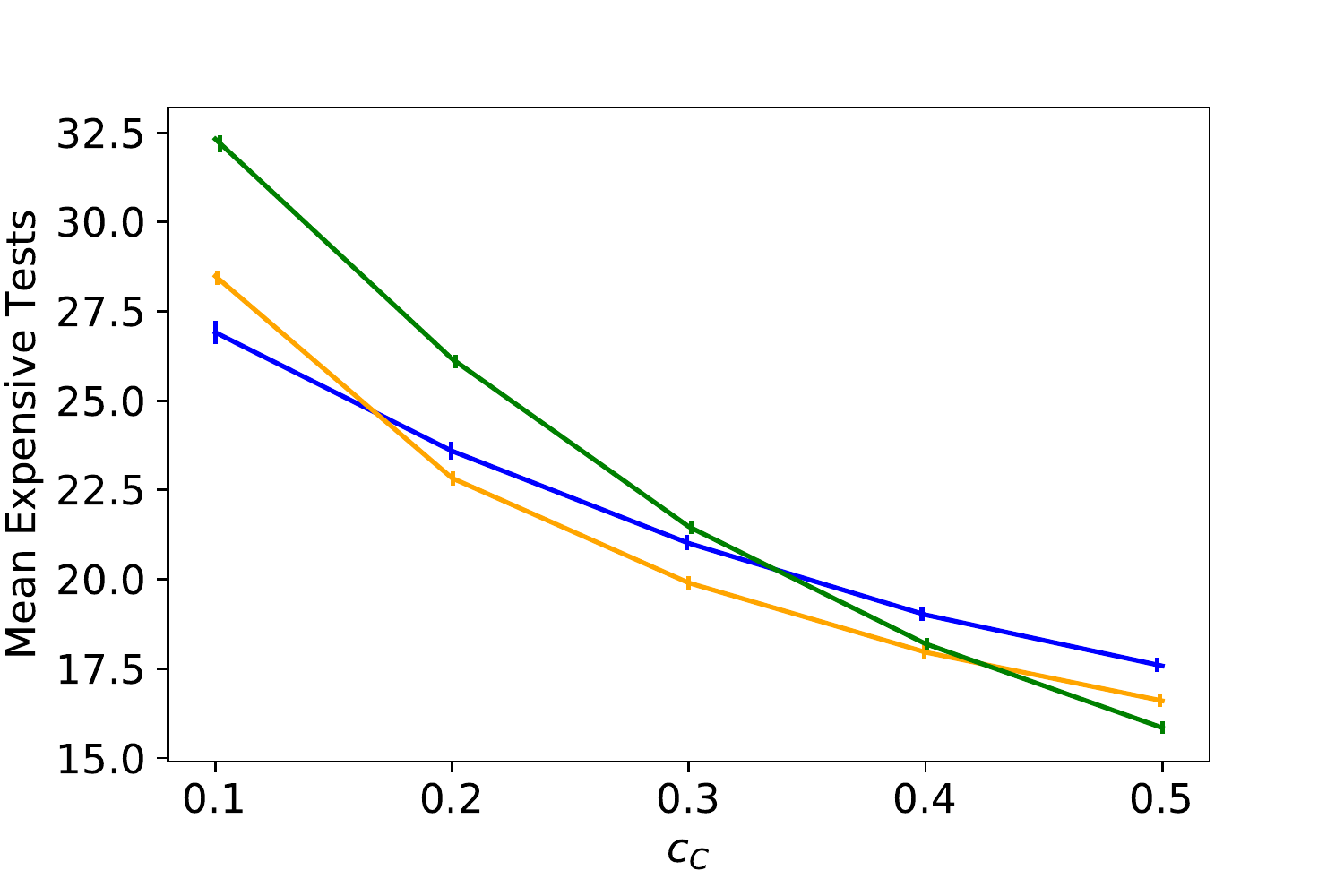}} 

\subfigure[]{\includegraphics[scale=0.4]{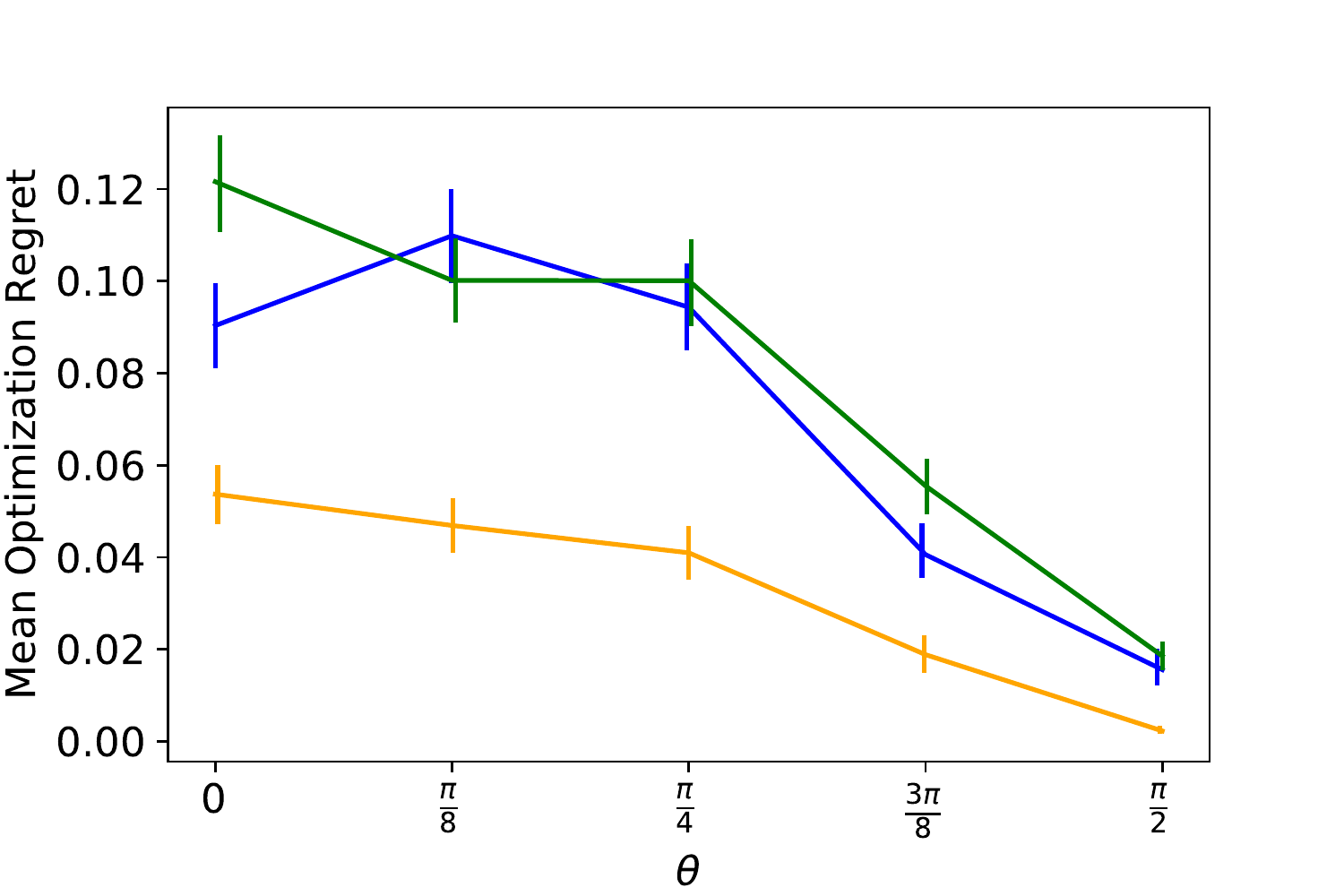}}\subfigure[]{\includegraphics[scale=0.4]{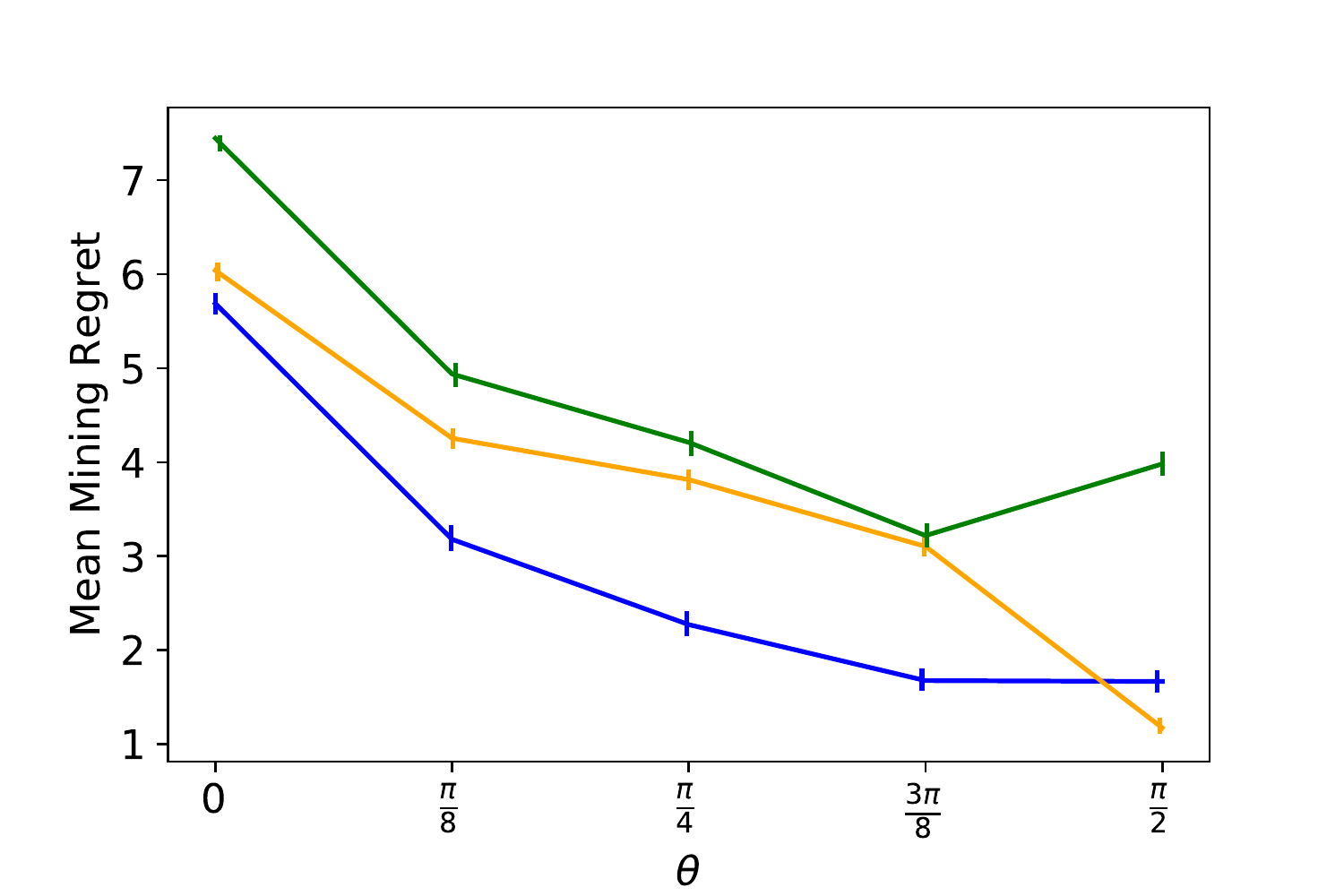}}\subfigure[]{\includegraphics[scale=0.4]{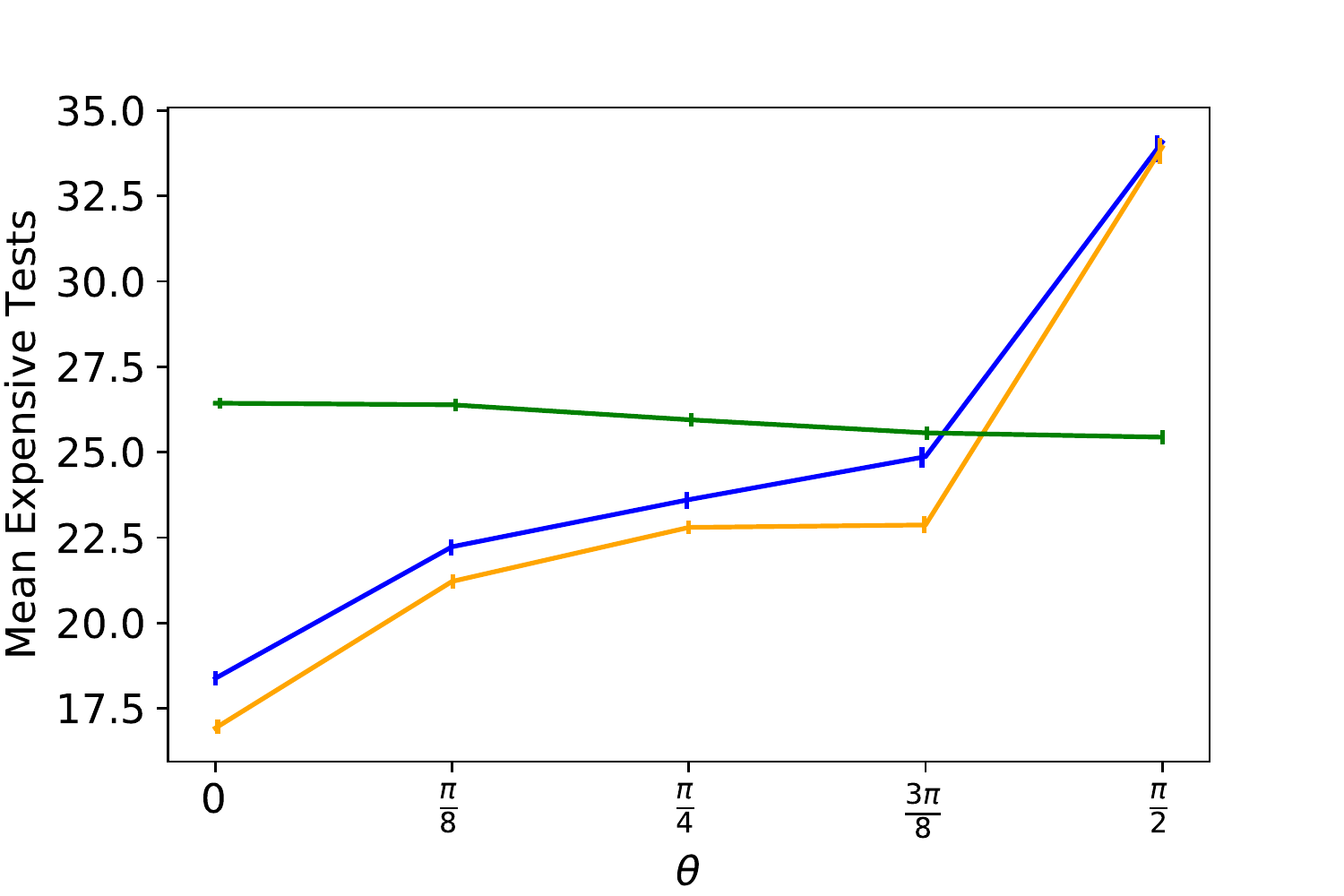}} 

\subfigure[]{\includegraphics[scale=0.4]{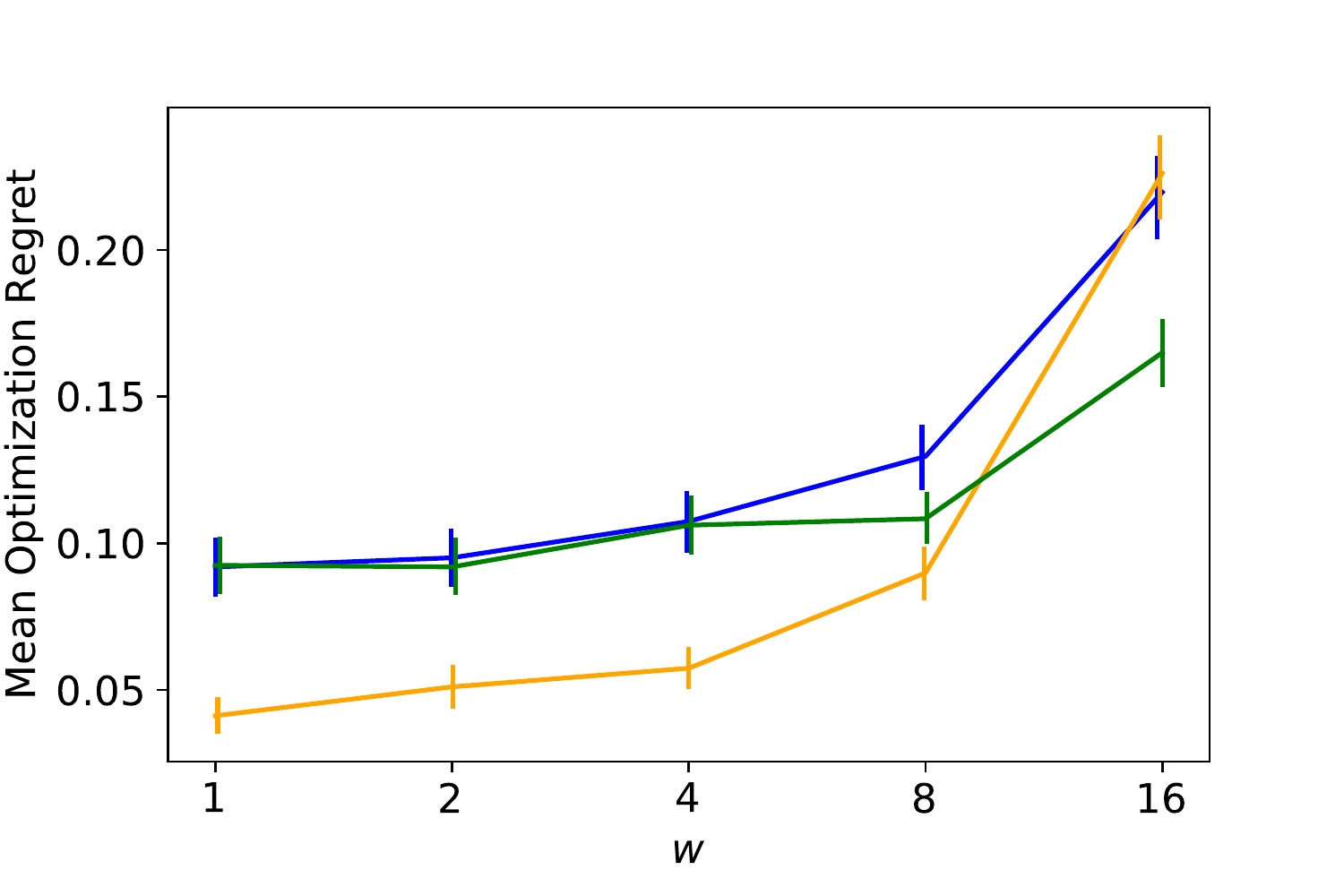}}\subfigure[]{\includegraphics[scale=0.4]{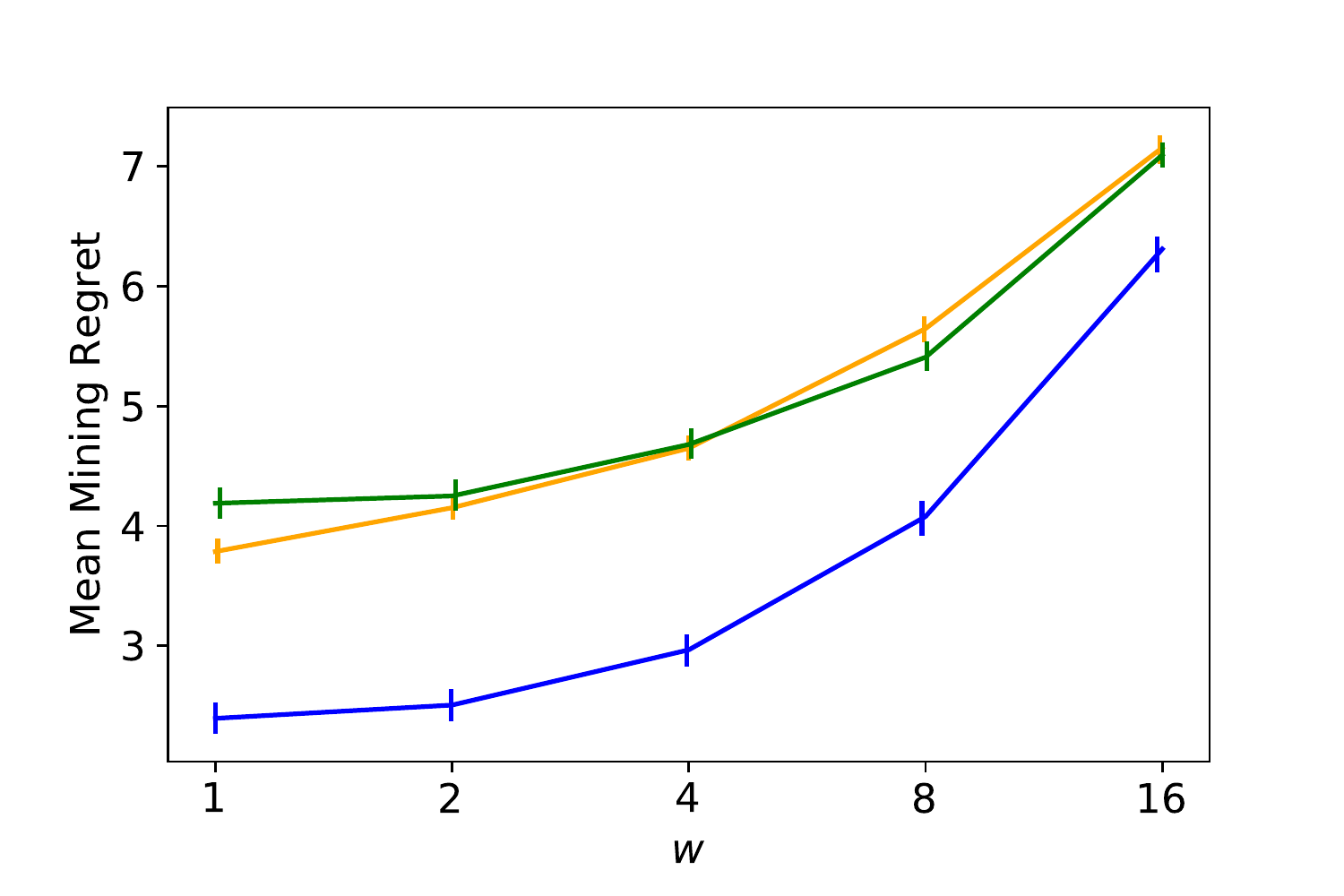}}\subfigure[]{\includegraphics[scale=0.4]{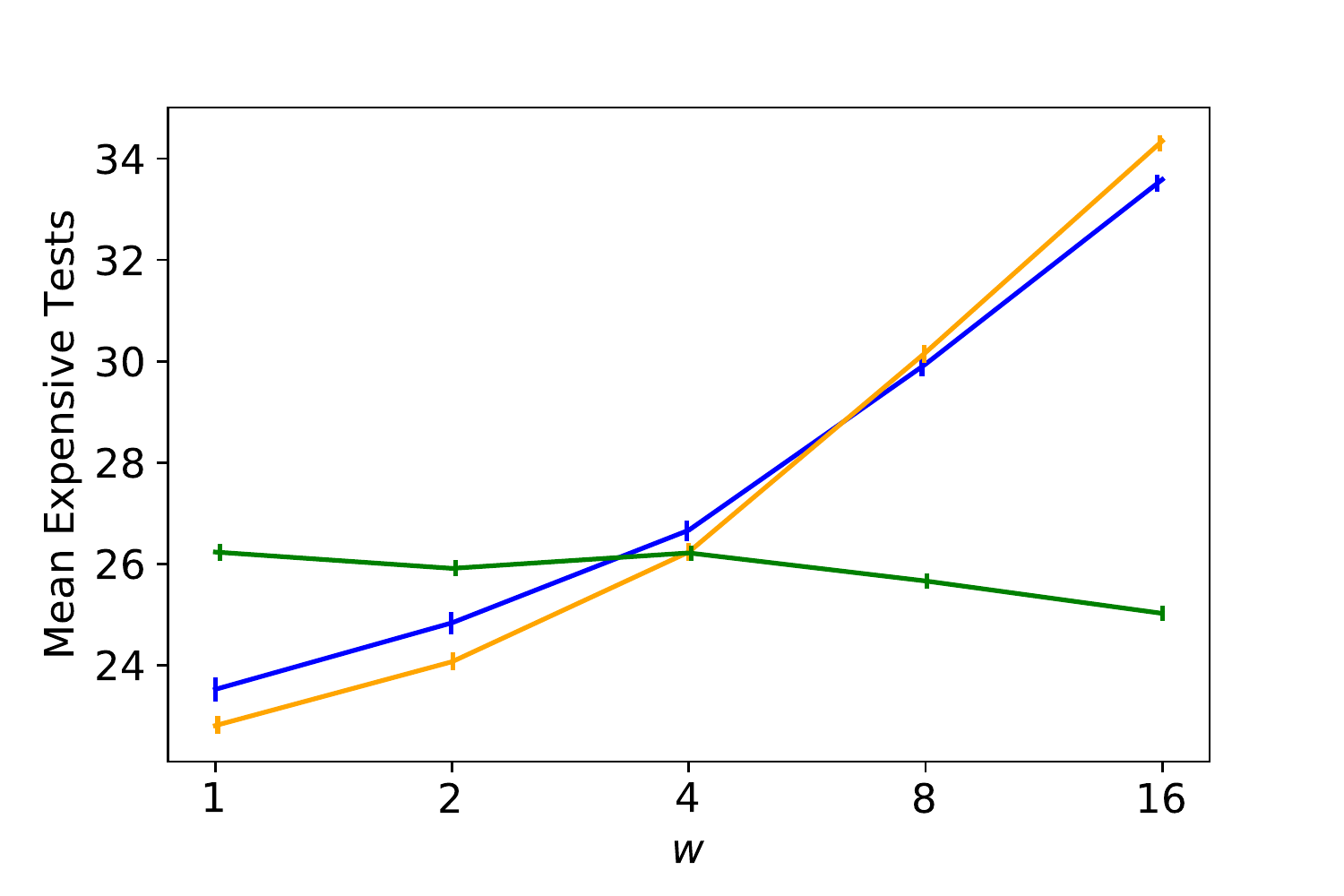}} 


\caption{Comparison of algorithm performance in screens on synthetic data described in Section~\ref{resultssynth}. (a,b,c) Experiment 1, (d,e,f) Experiment 2  and (g,h,i) Experiment 3. Bars indicate standard error of mean estimates. }\label{synthfig}
\end{center}
\end{figure*}

\begin{figure}[b!]
\begin{center}
\includegraphics[scale=0.4]{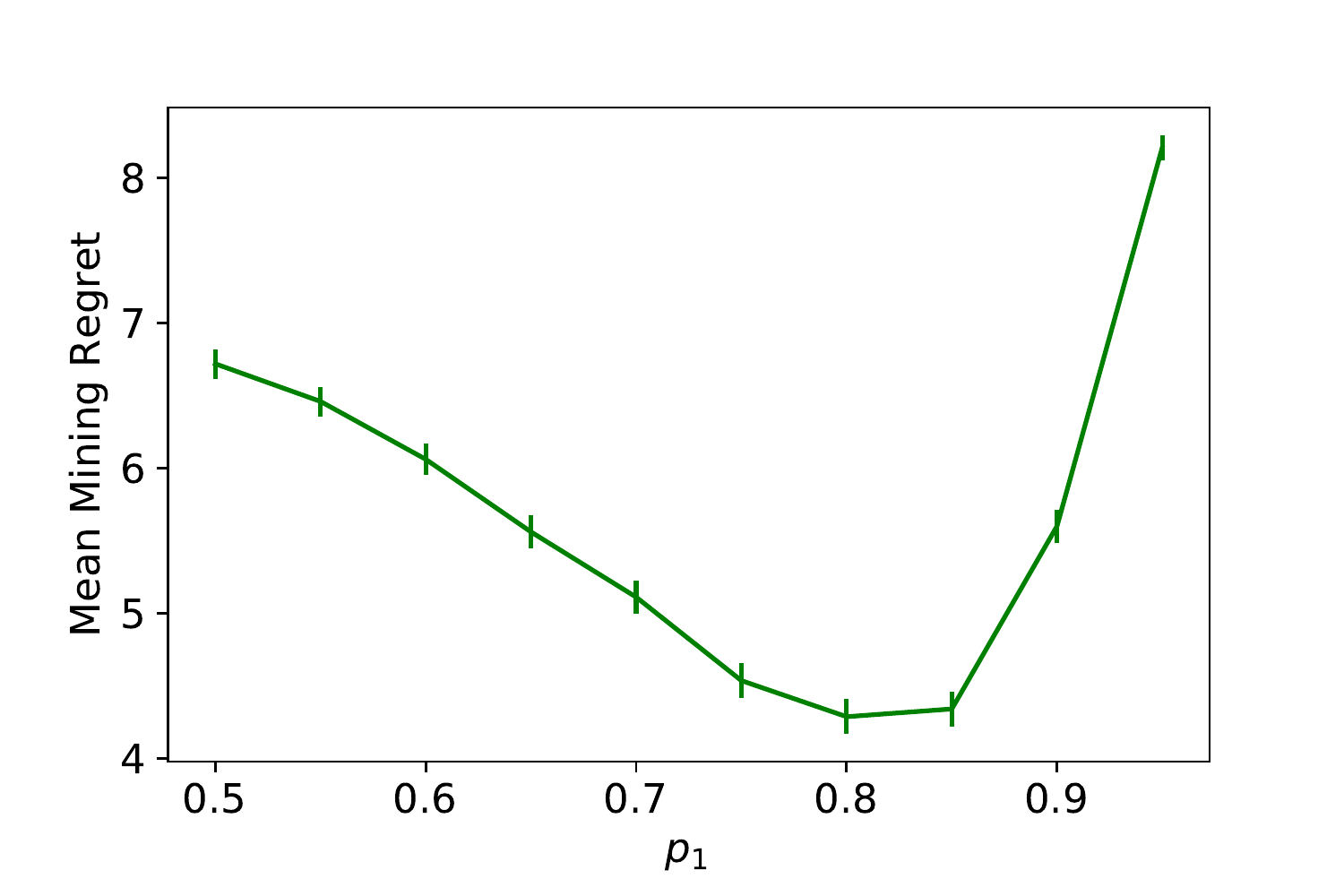} 
\caption{Varying $p_{1}$ parameter in Sequential Thompson Random method as described in Section~\ref{resultssynth}. Bars indicate standard error of mean estimates. }\label{thp1}
\end{center}
\end{figure}

\subsection{Results}\label{resultssynth}

We fix $c_{E}=1$ and $B=50$ and will vary $\theta$, $c_{C}$ and $w$.

\begin{enumerate}
    \item We fix $c_{C}=0.2$ and $w=1$ and vary $0\leq \theta\leq \pi/2$.
    \item We fix $w=1$, $\theta=\pi/4$ and vary $0.1\leq c_{C}\leq 0.5$.
    \item We fix $\theta=\pi/4$, $c_{C}=1$ and vary $1\leq w \leq 16$.
\end{enumerate}

For each set of experiment parameters we repeat 1000 independent trials and record the average optimization regret, average mining regret and the average number of expensive tests carried out. See Figure~\ref{synthfig}.

Note that Sequential Greedy Expected Improvement (SGEI), which targets the optimization objective, achieves the lowest average optimization regret for nearly all of the experiment parameter vales and likewise Sequential Greedy Threshold (SGT), which targets the mining objective,  achieves the lowest average mining regret. Sequential Thompson Random (STR) is outperformed by both Greedy methods in both metrics except for optimization regret when using the largest number of parallel workers.

We also experimented with varying $p_{1}$ in the STR method using $c_{C}=0.2$, $\theta=\pi/4$ and $w=1$. See Figure~\ref{thp1}. Note that although the optimally tuned method outperforms the method tuned using \eqref{thompsontune}, it is still some way behind SGT and that optimally tuning a parameter in this way would be impossible in a practical problem.

\section{Simulated screen on real data}

In this section we test our algorithms in simulated screens using real data from \emph{in silico} chemical engineering experiments. Metal Organic Frameworks (MOFs) \cite{Gen_MOF_Ref} and Covalent Organic Frameworks (COFs) \cite{Gen_COF_Ref} are families of porous solids that can be grown from a variety of component molecules into a vast array of different structures. Both MOFs and COFs have extremely high surface area to volume ratios which mean they can interact with gasses in special ways and have potential applications in a number of important industrial processes.


 We use the covariate testing model and the mining objective with $N=100$ so that the aim of the screen is to find as many of the top-100 materials as possible. We will compare our two-stage sampling algorithms with the following baseline methods: \newline

\noindent {\bf Single-Test Poor.} In this method we ignore the cheap test and apply standard single-test Bayesian optimization using the expensive test with the feature matrix $X$. \newline

\noindent  {\bf Single-Test Rich.} In this method we apply the cheap test to all of the candidates then apply standard single-test Bayesian optimization using the expensive test with the expanded feature matrix $[X,\bm{y}^{(E)}]$. \newline

In all of these experiments we fit the Gaussian process modelling hyperparameters to minimize the NLL of the data available at each step. Our code is based around the GPy package \cite{gpy2014} but with some custom modifications to to implement the covariate testing model and for sampling from large scale posteriors. See supplementary material for details.

\subsection{Methane deliverable capacity in COFs}

The Hypothetical Covalent Organic Framework (HCOF) database contains 69,839 material structures \cite{Mercado2018}. Each structure is provided with a vector of features that describe its composition and shape as well as its results in a simulated methane adsorption experiment. 

We choose methane deliverable capacity as the expensive test score which we want to target in the screen. Deliverable capacity in methane ($CH_4$) is defined by
\begin{equation}
\hbox{DC}_{CH_4}=\hbox{N}_{CH_4}^{{65 bar}}-\hbox{N}_{CH_4}^{{5.8 bar}}
\end{equation}
where $\hbox{N}_{CH_4}^{{p}}$ denotes the volume of methane at STP adsorbed by one unit volume of the material at pressure $p$. Roughly speaking this measures how efficiently the COF can store and release methane. We use 7 chemical composition features along with density as the 8 basic features in our screening problem. We use the void fraction for the cheap test  score. This measures what fraction of the structure's volume is open to the gas molecules. Note that void fraction is very important in determining deliverable capacity but via a non-linear relationship. See Figure~\ref{cofmoffig}. We use indicative costs of $c_{C}=0.1$, $c_{E}=1$ and allow a testing budget of $B=1000$. 

\begin{table*}[h]
\caption{Results for COF screen. All values averaged over 10 independent trials with standard deviation in brackets.}\label{coftab} 
\centering 
\begin{tabular}{c|c|cccc} 
\hline\hline 
method & workers & cheap tests & expensive tests & total cost & total reward \\ [0.5ex]
\hline 
GT - Rich & 1 & 69839 & 1000 & 70839 & 79.5 ~(2.6) \\ 
GT - Poor & 1 & 0 & 1000 & 1000 & 42.3 ~(4.3) \\ 
SGT & 1 & 2742.2 ~(247.7) & 725.1 ~(24.7) & 1000 & 63.9 ~(9.3) \\ 

\hline 
T - Rich & 100 & 69839 & 1000 & 70839 & 76.7 ~(3.2) \\ 
T - Poor & 100 & 0 & 1000 & 1000 & 39.0 ~(5.6) \\ 
SRT & 100 & 4901.1~(105.7) & 509.2~(10.6) & 1000 & 56.9 ~(2.9) \\  

\hline 
\end{tabular}
\label{tab:hresult}
\end{table*}

\begin{table*}[h]
\caption{Results for MOF screen. All values averaged over 10 independent trials with standard deviation in brackets.}\label{moftab} 
\centering 
\begin{tabular}{c|c|cccc} 
\hline\hline 
method & workers & cheap tests & expensive tests & total cost & total reward \\ [0.5ex]
\hline 
GT - Rich & 1 & 137953 & 1000 & 138953 & 98.0 ~(0.0) \\ 
GT - Poor & 1 & 0 & 1000 & 1000 & 70.1 ~(1.2) \\ 
SGT & 1 & 1613.4~(26.8) & 386.6~(26.8) & 1000 & 81.3 ~(2.2) \\ 

\hline 
T - Rich & 100 & 137953 & 1000 & 138953 & 100.0 ~(0.0) \\ 
T - Poor & 100 & 0 & 1000 & 1000 & 44.5 ~(6.8) \\ 
SRT & 100 & 1473.1~(18.9) & 526.9~(18.9) & 1000 & 67.1 ~(7.7) \\  

\hline 
\end{tabular}
\label{tab:hresult}
\end{table*}

\begin{figure}[t!]
 \begin{center}

 \subfigure[]{\includegraphics[scale=0.4]{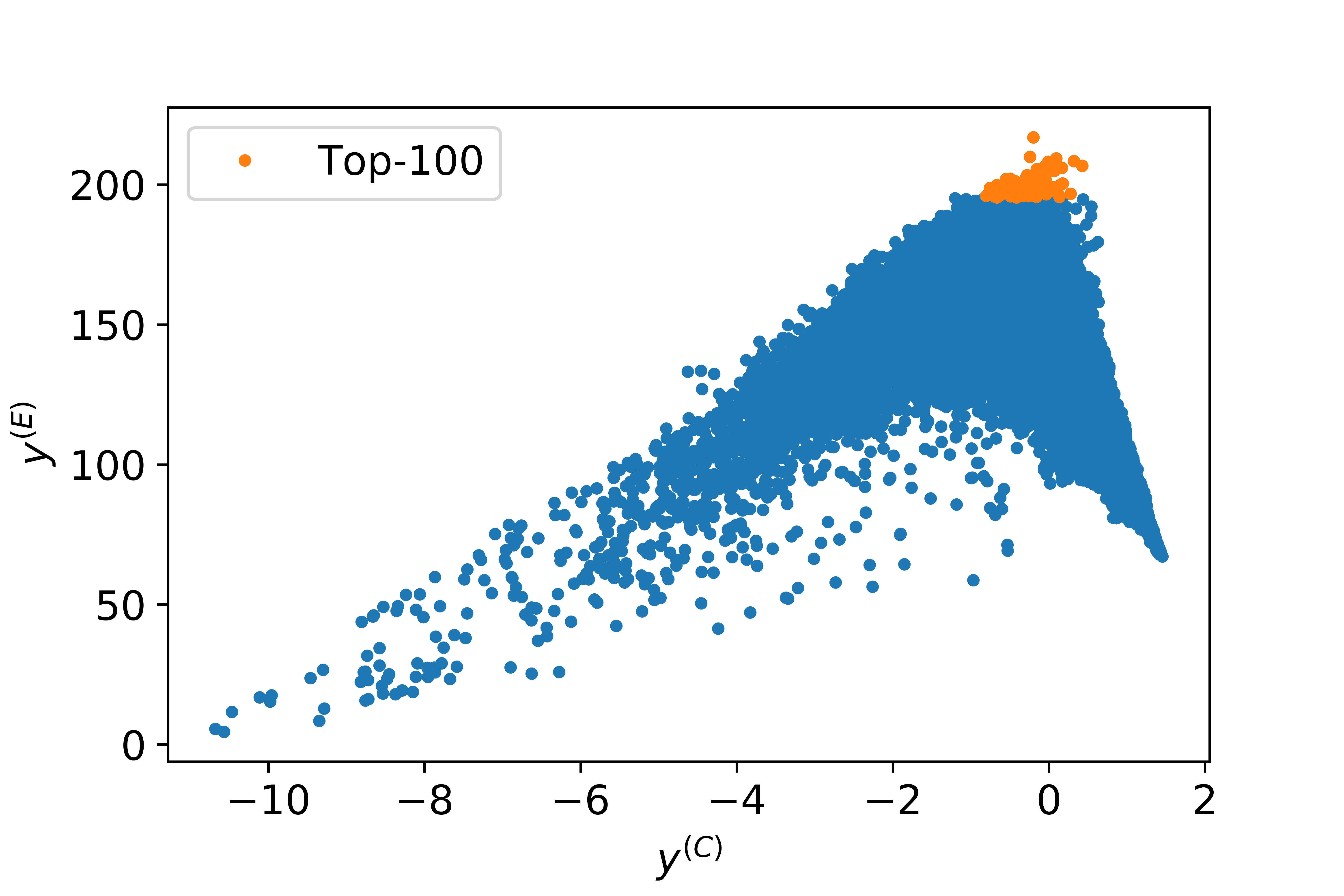}} \subfigure[]{\includegraphics[scale=0.4]{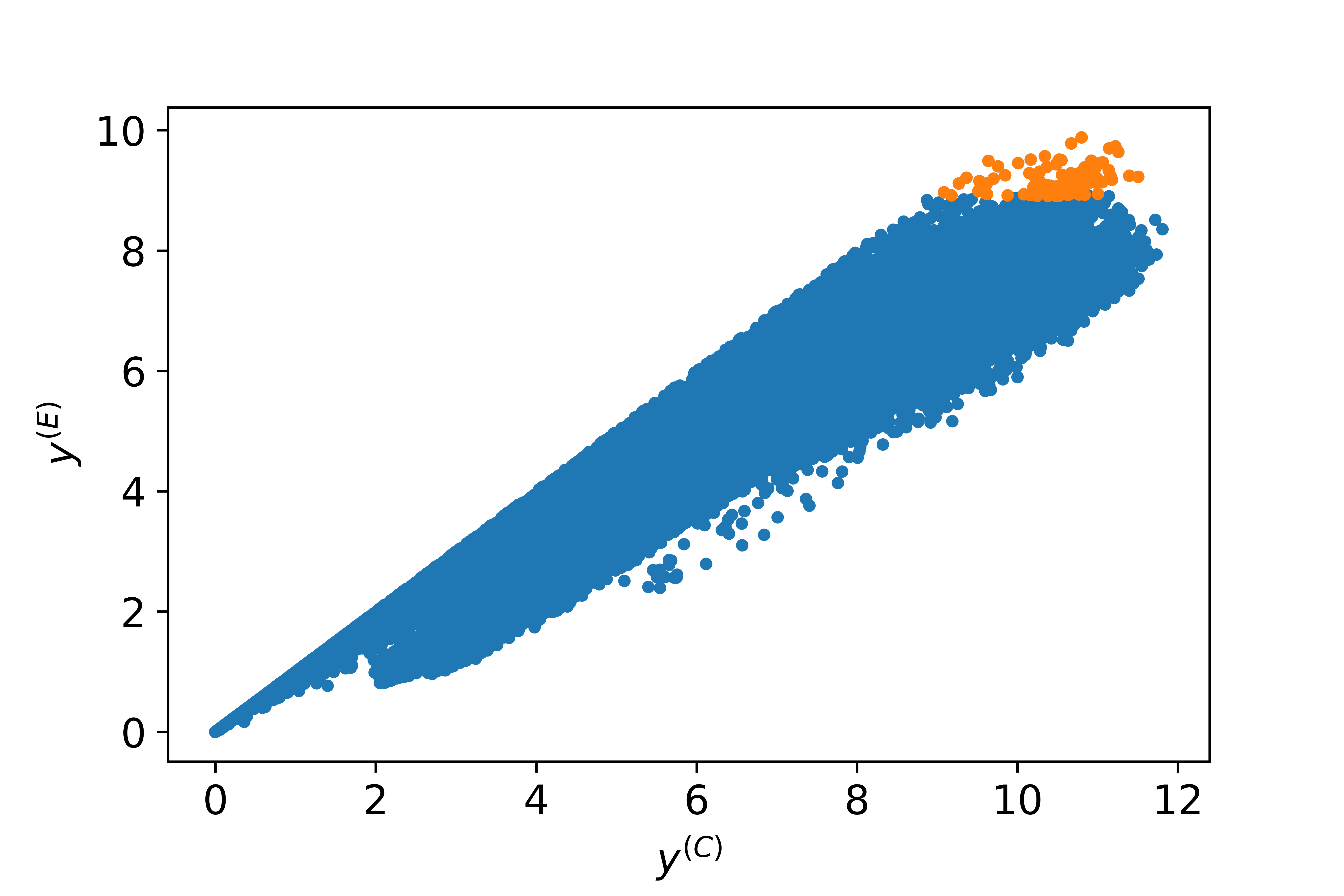}}
 
\caption{Test data used in screening problems. a) COF, b) MOFs.}\label{cofmoffig}
\end{center}
\end{figure}


 

\begin{figure*}[t!]
\begin{center}
\subfigure[]{\includegraphics[scale=0.4]{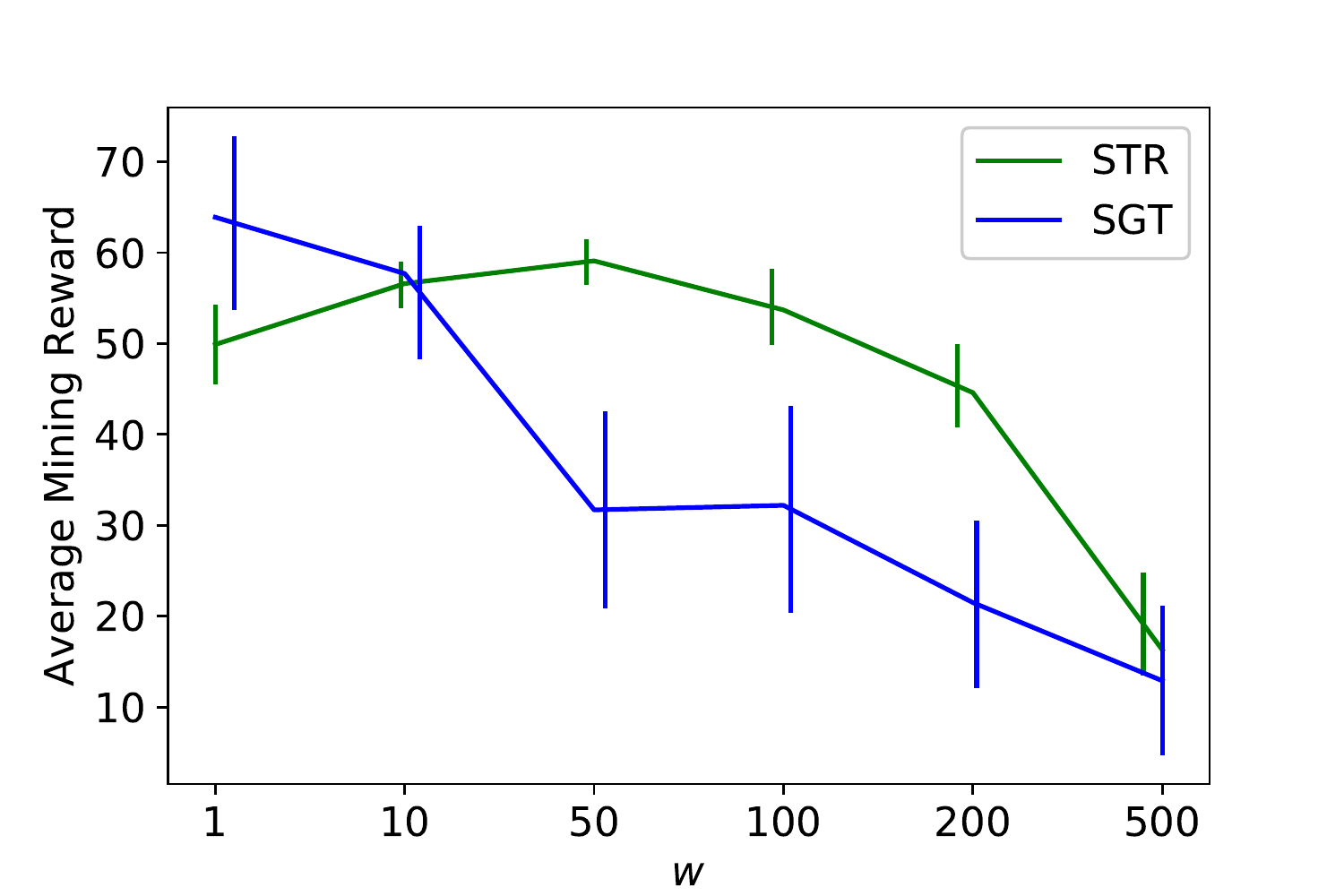}}\subfigure[]{\includegraphics[scale=0.03]{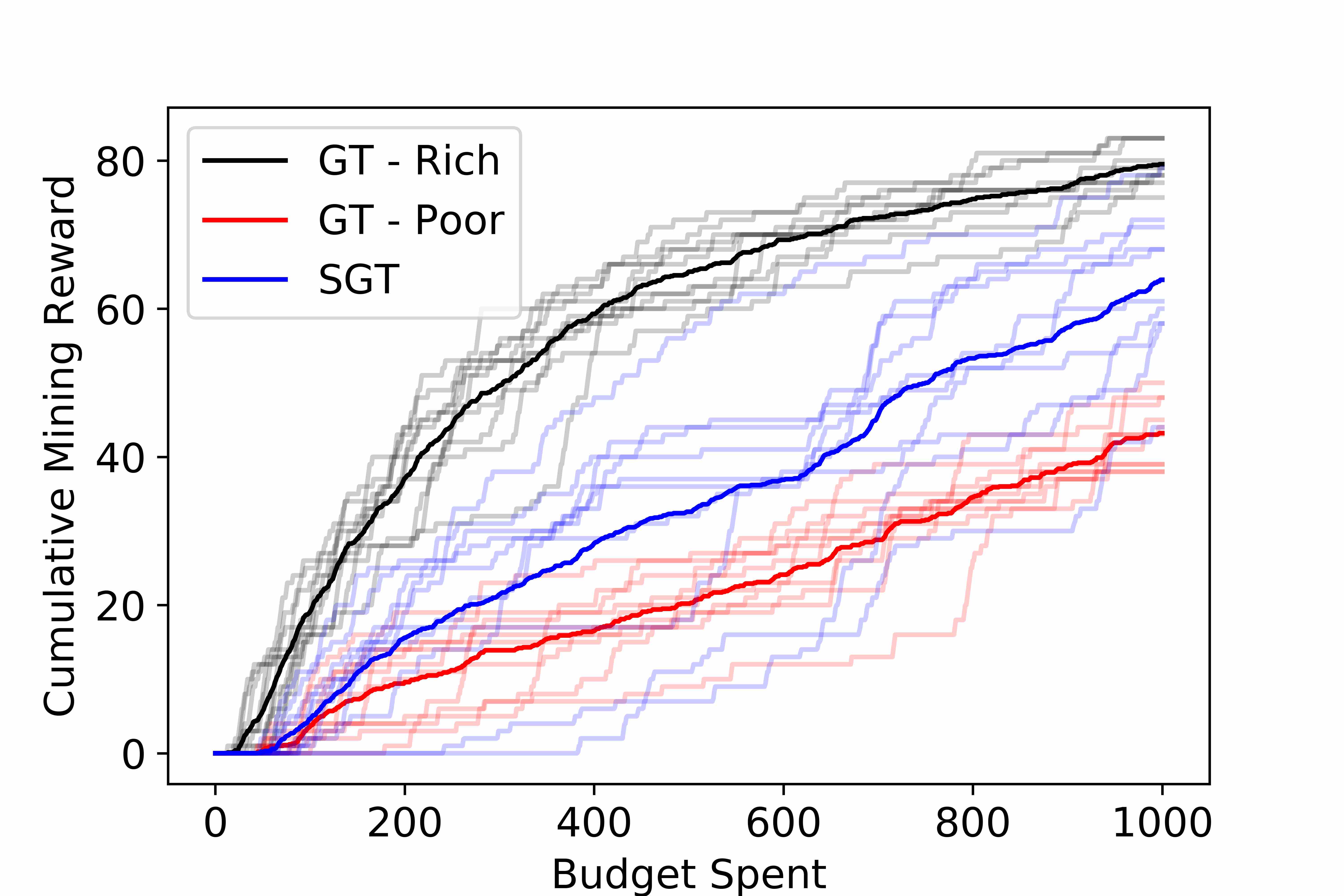}}\subfigure[]{\includegraphics[scale=0.03]{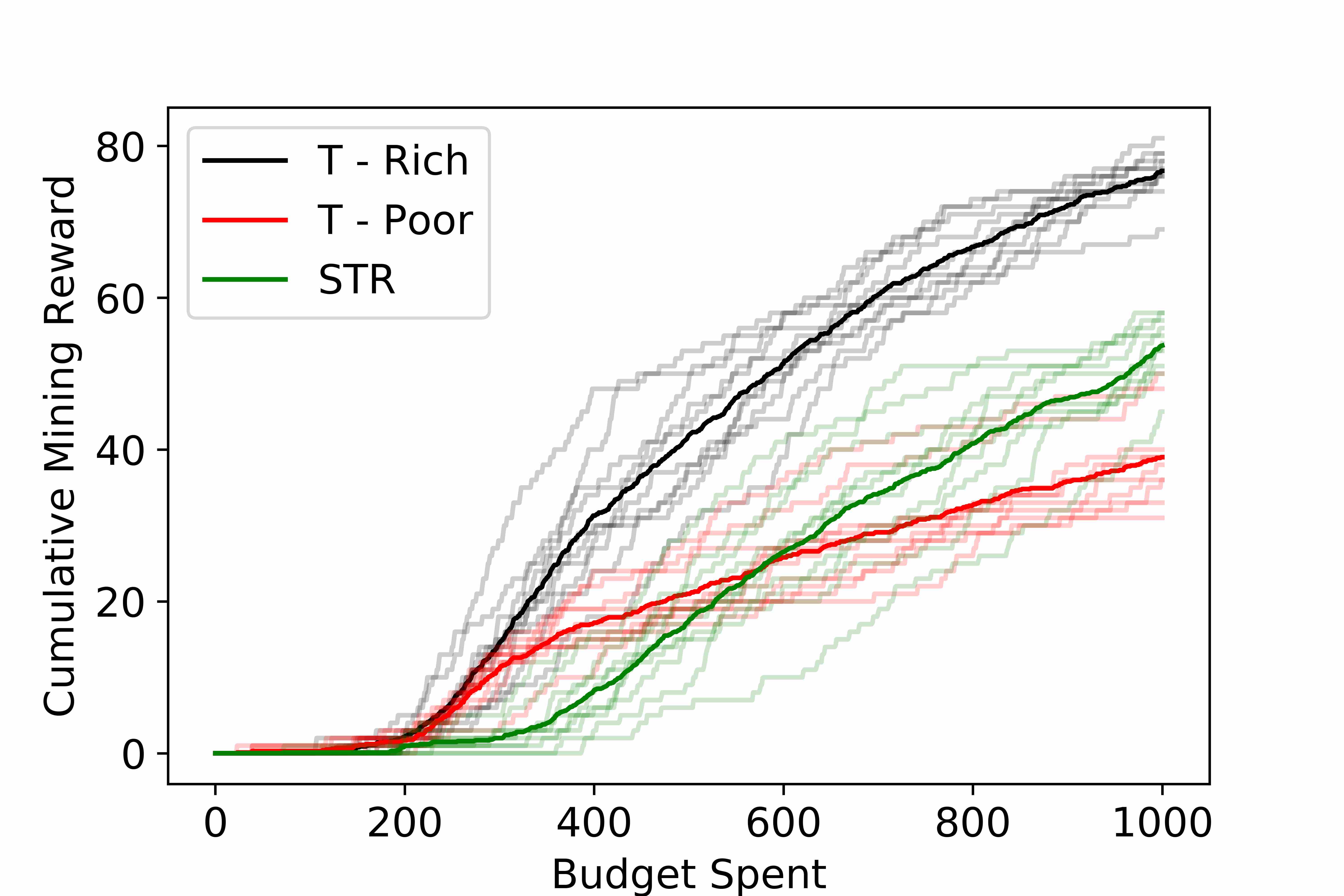}}

\subfigure[]{\includegraphics[scale=0.4]{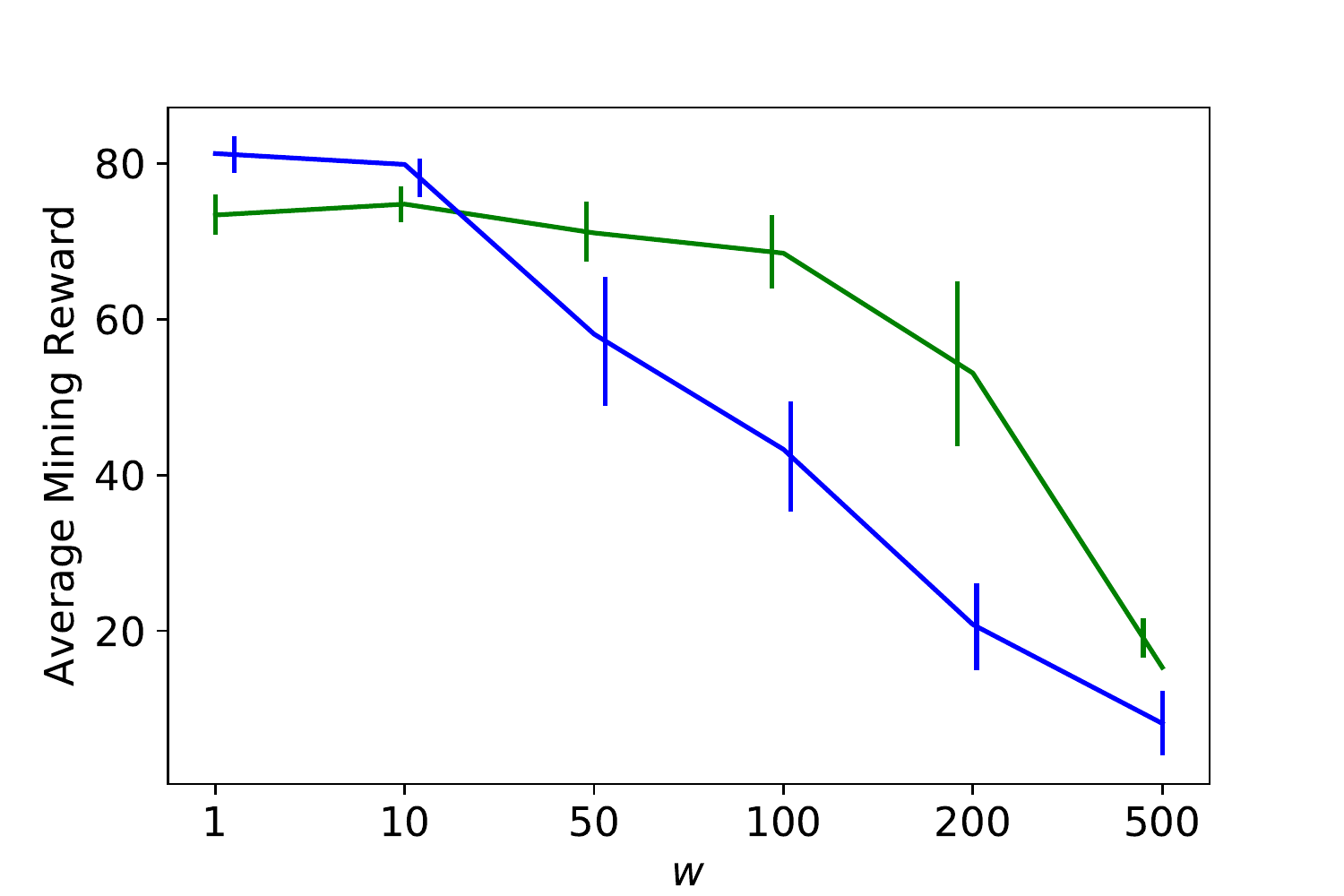}}\subfigure[]{\includegraphics[scale=0.03]{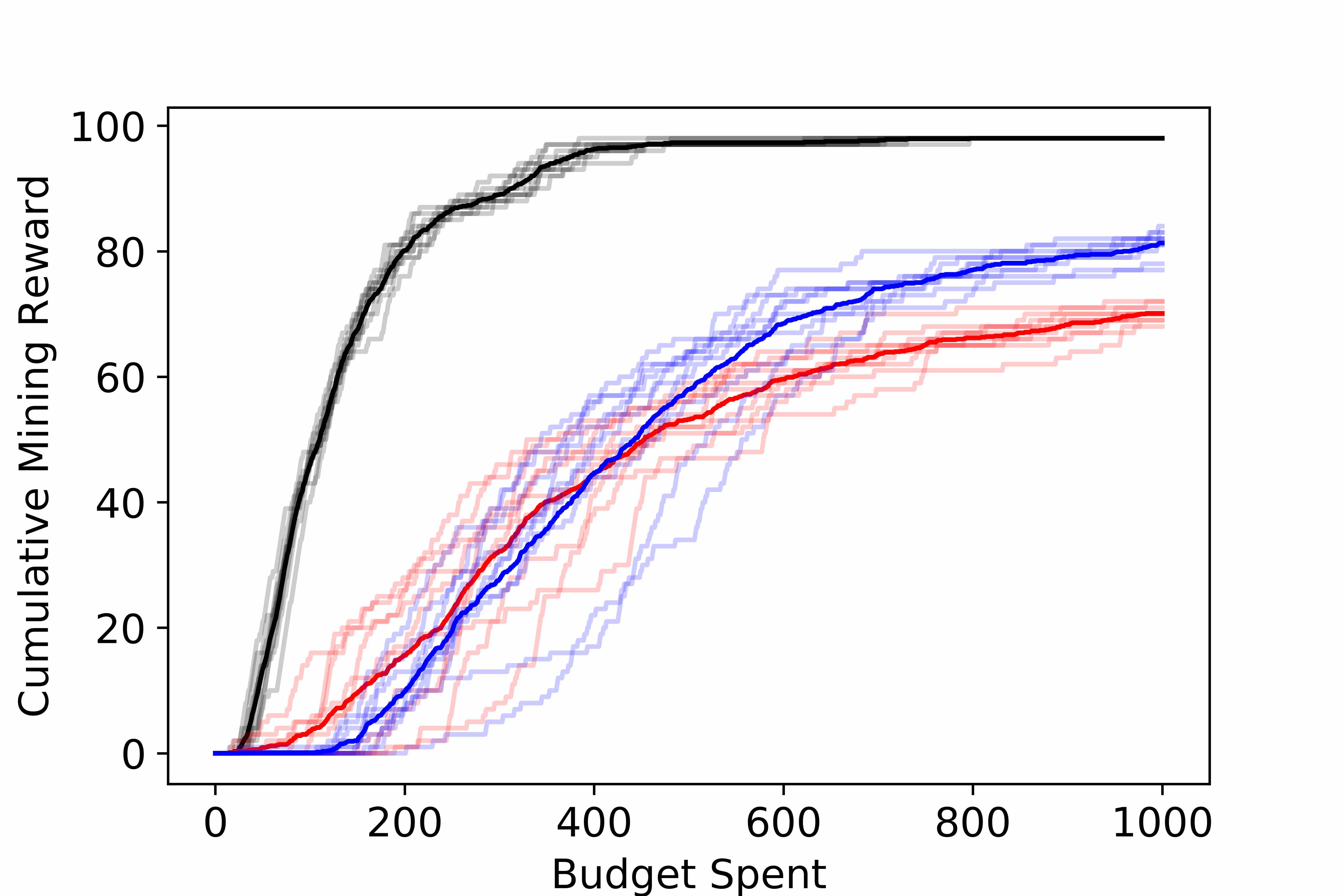}}\subfigure[]{\includegraphics[scale=0.03]{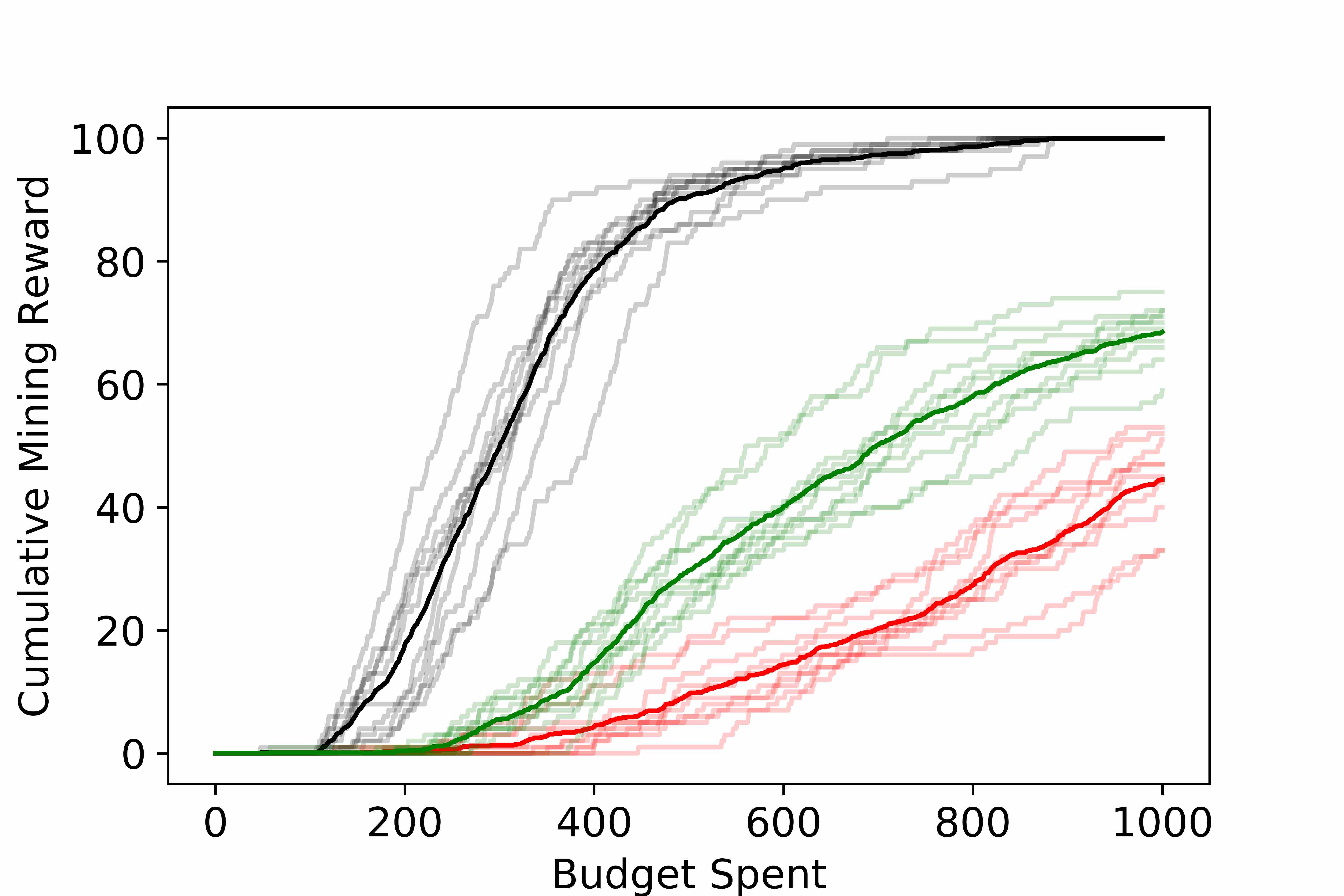}}



\caption{Comparison of algorithm performance in screens on (a,b,c) COF and (d,e,f) MOF data. (a,d) compare sequential methods over a range of worker numbers. (b,e) compare greedy methods with a single worker. (c,f) compare Thompson methods with 100 workers. Bars in (a,d) indicate standard deviation. Bold curves in (b,c,e,f) indicate averaged performance.}\label{cofmoffig} 
\end{center}
\end{figure*}

\subsection{Nitrogen Carbon dioxide separation in MOFs}

The HMOF database contains 137,953 material structures \cite{Wilmer2012a}. Each structure is provided with a vector of features that describe its composition and shape. We use 12 chemical composition features along with 6 physical features as the 18 features in our screening problem. We also use data from \cite{Fernandez2013} where each structure was tested for its ability to separate nitrogen and carbon dioxide. Rather than simulating a true mixture of the two gasses, which is far more computationally expensive, the authors simulated each gas adsorption separately and then combined the results of the two experiments to give an Adsorption Performance Indicator (API).
\begin{equation}
API=\frac{\hbox{DC}_{CO2}\times \hbox{N}_{CO2}^{{0.1 bar}}}{\hbox{N}_{N2}^{{0.9 bar}}+1}.
\end{equation}
In our screening problem we will use 
\begin{equation}
\bm{y}^{(C)}=\log\big(\hbox{DC}_{CO2}\times \hbox{N}_{CO2}^{{0.1 bar}}\big), \quad \bm{y}^{(E)}=\log\big(API\big),
\end{equation}
so that the cheap test scores are the contribution to the API from the CO$_{2}$ experiment and the expensive test scores are the final API. Note that we have taken the log so that the data is better suited to our Gaussian Process model. Also note that in this example the sequential testing restriction is naturally implied by the fact that the cheap test forms part of the expensive test. See Figure~\ref{cofmoffig}. We use indicative costs $c_{C}=0.5$, $c_{E}=0.5$ and allow a testing budget of $B=1000$.  When scoring single-test methods on this problem we use $c_{E}=1$ for a fair comparison. 

\subsection{Results}

We simulated the screening process using the Sequential Greedy Threshold (SGT) and Sequential Thompson Random (STR) two-test methods and compared them to the Greedy Threshold (GT) and Thompson (T) single-test methods with the Rich and Poor datasets. Each experiment was repeated over ten independent trials. 

Figure~\ref{cofmoffig} (a,d) compares the performance of SGT and STR for the two databases. Note that for a single worker the SGT method is most efficient but that for fifty or more workers the STR method performs best. In the COF experiment we actually observed improved performance with increased number of workers for $1\leq w\leq 50$, which may be due to the increased number of (uniformly) randomly chosen initial samples taken with more workers. 

Figure~\ref{cofmoffig} (b,c,e,f) compares the performance of our chosen two-test method to the corresponding single-test method using SGT with one worker or STR with one hundred workers. The results of these experiments are also summarised in Tables~\ref{coftab} and~\ref{moftab}. In both cases the two-test method is able to significantly outperform the single-test method on the poor dataset (which is its only fair comparison). 

\section{Conclusions}

We presented new models and algorithms for multi-test Bayesian optimization for application in large scale materials screening problems. We have demonstrated the potential power of these techniques in two simulated screens on real data from previous screening studies. Using the covariate testing model allows our algorithms to learn complex non-linear patterns but makes computation difficult. 

Some possible direction for future algorithm development include the following.

\noindent {\bf Non-sequential methods.} As discussed the sequential condition to always apply the cheap test before the expensive test could be inefficient in some problems. However making inferences when $I_{\hbox{ut}}$ is non-empty is a difficult missing data problem that will very challenging in large-scale moderate-dimensional problems.

\noindent {\bf Entropy based methods.} We do not currently have an efficient way to adapt the entropy acquisition function to the multi-test setting. We did experiment with an entropy based controller for use with Thompson sampling but this was too slow to even carry out a large number of small scale experiments. 

\noindent {\bf More than two tests.} The covariate testing model could be adapted to support a range of different cheap tests, possibly all related to the expensive test scores by very different non-linear relationships. Any efficient sampling method that worked with this model would need to work in a non-sequential manner as described above.

\section{Acknowledgements}
This work was supported by the Engineering and Physical Sciences Research Council EP/L016354/1.
This project has received funding from the European Research Council (ERC) under the European Union's Horizon 2020 research and innovation programme (grant agreement No 648283 GROWMOF)
This research made use of the Balena High Performance Computing (HPC) Service at the University of Bath.

\section{Supporting Information}
The code used as part of this study can be found at:
\begin{verbatim}
    https://gitlab.com/AMInvestigator/ame/-/tree/multi_test_bayesian_optimisation
\end{verbatim}


\bibliography{main.bib}
\bibliographystyle{plain}



\end{document}